\documentclass{article}
\usepackage[utf8]{inputenc}
\usepackage{graphics}
\usepackage{tikz}
\usetikzlibrary{arrows,shapes,positioning}

\usepackage{amsmath}
\usepackage{amsthm}
\usepackage{amssymb}

\newtheorem{definition}{Definition}

\usepackage{natbib}
\usepackage{url}

\usepackage{tikz}
\usepackage{pgfplots}

\usepackage{multirow}

\title{Declarative Statistics\thanks{This work integrates and extends the discussion in \citep{rossi2014}.}}
\author{Roberto Rossi,${}^1$ \"Ozg\"ur Akg\"un,${}^2$ Steven Prestwich,${}^3$\\
S. Armagan Tarim${}^4$\\
\\
${}^1$Business School, University of Edinburgh, UK\\
${}^2$Department of Computer Science, University of Saint Andrews, UK\\
${}^3$Insight Centre for Data Analytics, University College Cork, Ireland\\
${}^4$Department of Management, Cankaya University, Turkey
}

\date{}

\begin{document}
\maketitle

\begin{abstract}
In this work we introduce declarative statistics, a suite of declarative modelling tools for statistical analysis. Statistical constraints represent the key building block of declarative statistics. First, we introduce a range of relevant counting and matrix constraints and associated decompositions, some of which novel, that are instrumental in the design of statistical constraints. Second, we introduce a selection of novel statistical constraints and associated decompositions, which constitute a self-contained toolbox that can be used to tackle a wide range of problems typically encountered by statisticians. Finally, we deploy these statistical constraints to a wide range of application areas drawn from classical statistics and we contrast our framework against established practices.
\end{abstract}

\section{Introduction}\label{sec:intro}

In this work we develop declarative modelling tools for statistical analysis; we name the resulting modeling paradigm {\em declarative statistics}.

Declarative statistics provides a platform to model a family of hypothesis about a phenomenon being studied and to determine which of these hypothesis is ``compatible'' with available data. A feasible assignment in a declarative statistics model represents one of the many possible hypothesis that are ``compatible'' with available data, i.e. that --- to use statistical terminology --- we failed to reject. Conversely, infeasible assignments represent hypothesis that have been rejected, in a statistical sense, at the prescribed significance level on the basis of the observations that are available. In short, to paraphrase \cite{citeulike:12517605}, declarative statistics recognises that ``there are no Models, there are only models;'' and that data are the ultimate discriminant that determines which model we should retain and which one we should drop.

Statistical constraints \citep{rossi2014}, a recently introduced modeling paradigm that links statistics and constraint programming, represent the key building block of declarative statistics. Informally speaking, a statistical constraint exploits statistical inference to determine what decision variable assignments satisfy a given statistical property at a prescribed significance level. For instance, a statistical constraint may be used to determine, for a given distribution, what values for one or more of its parameters, e.g. the mean, are consistent with a given set of samples. Alternatively, it may be used to determine what sets of samples are compatible with one or more hypothetical distributions.

Although it is possible --- like in least squares regression --- to use a declarative statistics model to obtain ``best-fit values'' for problem parameters, this is not the main goal of declarative statistics. Declarative statistics aims to provide a representation of the confidence region associated with problem parameters; a representation which a decision maker can interact with via a high level modeling framework. 

\begin{figure}[h!]
\centering
\includegraphics{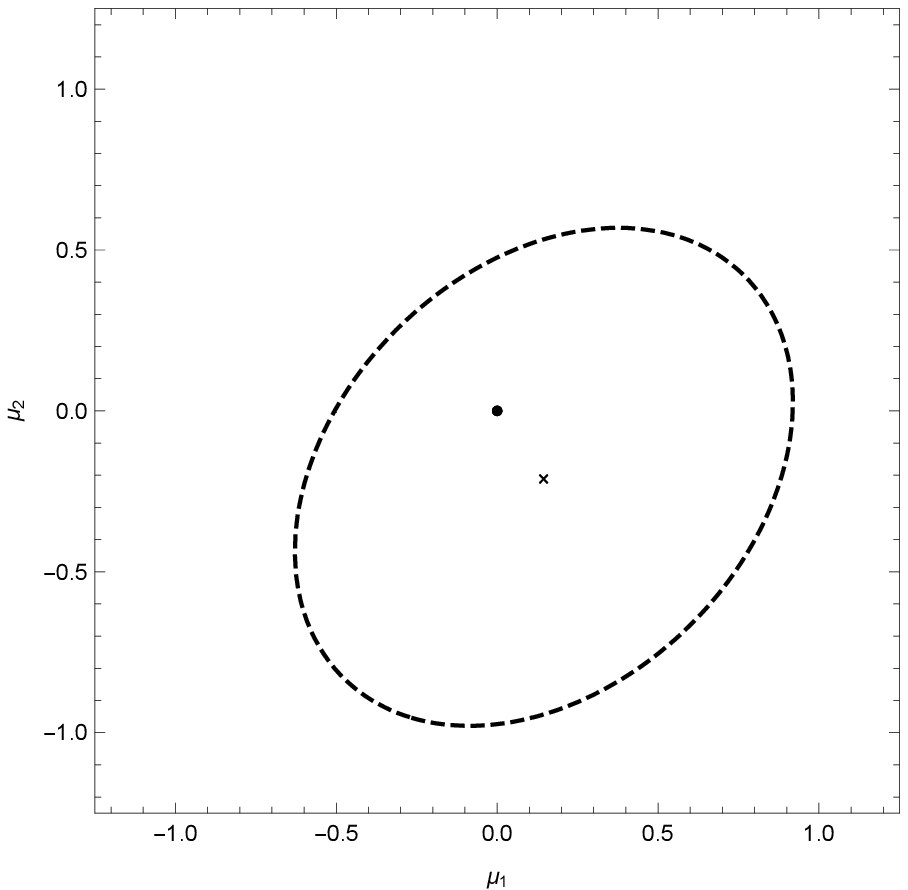}
\caption{Confidence region for the location parameter $\mu\in \mathbb{R}^2$ of a bivariate normal distribution.}
\label{fig:confidence_region_bivariate_normal}
\end{figure}

To focus attention, we draw an example from classical statistical analysis of a multivariate normal distribution. If provided with a set of $M$ samples drawn from a bivariate normal distribution, a statistician can use established results \citep[see e.g.][chap. 5]{citeulike:14214028} to construct a confidence region representing values of the two-dimensional mean vector $\mu$ that are likely to have generated the observed samples at the prescribed significance level $\alpha$. The region obtained looks like the one shown in Fig. \ref{fig:confidence_region_bivariate_normal}. In this picture, the black dot represents the ``true'' vector that generated the observations, the cross is the sample mean. A region computed from a different set of $M$ samples is likely to feature a different shape and to shift to a different part of the quadrant. However, if the statistician computes such region over and over again for different sets of $M$ samples, in the long run with probability $1-\alpha$ these regions will cover the ``true'' vector that generated the observations. 

Declarative statistics offers a high level modeling framework that can be used to represent confidence regions for problems like the one just illustrated. These regions can be queried in different ways: the decision maker may ask what vector in the region is the most likely one, or explore the boundaries of the region to construct confidence intervals for model parameters --- in the above example, confidence intervals for the two-dimensional mean vector $\mu$. Most importantly, declarative statistics provides a framework that can capture not only the confidence region of known problems like the one presented above, but of complex families of statistical hypothesis expressed via a high-level modeling framework. 

Having provided an informal high-level description of declarative statistics, by building upon the discussion in \citep{rossi2014}, in the rest of this work we make the following contributions to the literature.
\begin{itemize}
\item We introduce a range of relevant counting and matrix constraints and associated decompositions, some of which novel, to deal with challenges typically faced in the design of statistical constraints.
\item We introduce a selection of novel statistical constraints and associated decompositions, which constitute a self-contained toolbox that can be used to tackle a wide range of problems typically encountered by statisticians.
\item We deploy the aforementioned statistical constraints to a wide range of application areas drawn from classical statistics and we contrast our framework against established practices.
\end{itemize}
We believe our framework is more expressive than other existing solutions. In contrast to established approaches such as least squares regression, our strategy does not put excessive emphasis of specific statistics, such as mean values, but provides a comprehensive representation of all hypothesis that, within the family being scrutinised, are compatible with the data. If necessary, these hypothesis can be ranked by relying on given statistics, including --- but not necessarily --- statistics representing a euclidean distance like the mean squared error.

This work is structured as follows. In Section \ref{sec:formal_background} we introduce the relevant formal background in constraint programming, statistical inference, and statistical constraints. In Section \ref{sec:counting_matrix} we introduce relevant counting and matrix constraints that are instrumental to our analysis. In Section \ref{sec:selection_statistical_constraints} we present a selection of novel statistical constraints; and in Section \ref{sec:applications} we deploy these constraints in a wide range of practical applications drawn from classical statistics. In Section \ref{sec:related_works} we discuss related works; and in Section \ref{sec:conclusions} we draw conclusions.

\section{Formal background}\label{sec:formal_background}

In this section we introduce the relevant formal background in constraint programming, statistical inference, and statistical constraints, whose aim is to blend the former two subject areas.

\subsection{Constraint programming}

A Constraint Satisfaction Problem (CSP) is a triple $\langle V,C,D \rangle$, where $V$
is a set of decision variables, $D$ is a function mapping each element of $V$ to a domain of
potential values, and $C$ is a set of constraints stating allowed combinations of values for subsets of variables in $V$ \citep{1207782}. A {\em solution} to a CSP is an assignment of variables to values in their respective domains such that all of the constraints are satisfied. 

CP features several classes of constraint, e.g. logical constraints, linear constraints, and {\em global constraints} \citep{reg03}. A global constraint captures a relation among a non-fixed number of variables. 
Constraints embed dedicated {\em filtering algorithms} able to remove provably infeasible
or suboptimal values from the domains of the decision variables that are constrained and, therefore, to enforce some degree of {\em consistency}, e.g. arc consistency (AC) \citep{citeulike:6598311}, bounds consistency (BC) \citep{Choi06} or generalised arc consistency (GAC) \citep{Freuder:1982:SCB:322290.322292}. A constraint is \emph{generalized arc-consistent} if and only if, when a variable is assigned any of the values in its domain, there exist compatible values in the domains of all the other variables in the constraint. Filtering algorithms are repeatedly called until no more values are pruned; this process is called {\em constraint propagation}.

In addition to constraints and filtering algorithms, constraint solvers also feature a heuristic {\em search engine}, e.g. a backtracking algorithm guided by dedicated variable and value selection heuristics. During search, the constraint solver explores partial assignments and exploits filtering algorithms in order to proactively prune parts of the search space that cannot lead to a feasible or to an optimal solution.

\subsection{Constraints capturing a statistic}

Constraints capturing statistics such as mean and standard deviation have been long known and investigated in the constraint programming literature, see e.g. \citep{citeulike:13171963,citeulike:14181517}. In the context of this work, we shall draw a clear demarcation line between constraints capturing statistics, such as a constraint computing a mean or a standard deviation, and statistical constraints, such as the $t$-test constraint \citep{rossi2014}, which force a statistic to satisfy some condition.  

Let $X\equiv\{x_1,\ldots,x_n\}$ be decision variables, in the rest of this work we shall consider the following constraints capturing a statistic.

\begin{definition}
\textsc{mean}$(X,\mu)$ holds iif $\mu = \sum_{i=1}^{n}x_i/n$.
\end{definition}

\begin{definition}
\textsc{variance}$(X,\sigma^2)$ holds iif $\sigma^2 = \sum_{i=1}^{n}(x_i-\mu)^2/(n-1$), where \textsc{mean}$(X,\mu)$.
\end{definition}

\begin{definition}
\textsc{standard\_deviation}$(X,\sigma)$ iif $\sigma = \sqrt{\sum_{i=1}^{n}(x_i-\mu)^2/(n-1)}$, where \textsc{mean}$(X,\mu)$.
\end{definition}

\begin{definition}
\textsc{standard\_error}$(X,s)$ holds iif $s = \sigma/\sqrt{n}$, where we define \textsc{standard\_deviation}$(X,\sigma)$.
\end{definition}

Finally, let $Y\equiv\{y_1,\ldots,y_n\}$ be decision variables.

\begin{definition}
\textsc{covariance}$(X,Y,\sigma^2)$ iif $\sigma^2 = \sum_{i=1}^{n}(x_i-\mu_x)(y_i-\mu_y)/(n-1$), where \textsc{mean}$(X,\mu_x)$ and \textsc{mean}$(Y,\mu_y)$.
\end{definition}

Although several works have investigated effective filtering strategies for constraints capturing statistics, this investigation falls out of the scope of our work. 

In order to model statistics such as mean, standard deviation, variance and covariance, in the rest of this work we will rely on interval arithmetic extensions discussed in \citep{tricks2013}, which makes it possible to model real variables and constraints within a Choco\footnote{\url{http://www.choco-solver.org/}} \citep{choco} model and delegate associated reasoning during search to Ibex,\footnote{\url{http://www.ibex-lib.org/}} a library for constraint processing over real numbers.

\subsection{Statistical inference}\label{sec:inference}

A probability space, as introduced in \citep{kolmogorov1960foundations}, is a mathematical tool that aims at modelling a real-world experiment consisting of outcomes that occur randomly. As such it is described by a triple $(\Omega,\mathcal{F}, \mathcal{P})$, where $\Omega$ denotes the sample space --- i.e. the set of all possible outcomes of the experiment; $\mathcal{F}$ denotes the sigma-algebra on $\Omega$ --- i.e. the set of all possible events on the sample space, where an event is a set that includes zero or more outcomes; and $\mathcal{P}$ denotes the probability measure --- i.e. a function $\mathcal{P}:\mathcal{F}\rightarrow[0,1]$ returning the probability of each possible event. A random variable $\omega$ is an $\mathcal{F}$-measurable function $\omega:\Omega \rightarrow \mathbb{R}$ defined on a probability space $(\Omega,\mathcal{F}, \mathcal{P})$  mapping its sample space to the set of all real numbers. Given $\omega$, we can ask questions such as ``what is the probability that $\omega$ is less or equal to element $s\in\mathbb{R}$.'' This is the probability of event $\{o: \omega(o) \leq s \}\in\mathcal{F}$, which is often written as $F_{\omega}(s)=\Pr(\omega\leq s)$, where $F_{\omega}(s)$ is the cumulative distribution function (CDF) of $\omega$. A multivariate random variable is a random vector $(\omega_1,\ldots,\omega_n)^{T}$, where ${}^{T}$ denotes the ``transpose'' operator. If $\omega_1,\ldots,\omega_n$ are independent and identically distributed (iid) random variables, the random vector may be used to represent an experiment repeated $n$ times, i.e. a sample, where each replica $i$ generates a {\em random variate} $\omega'_i$ and the outcome of the experiment is vector $(\omega'_1,\ldots,\omega'_n)^{T}$. 

Consider a multivariate random variable defined on probability space $(\Omega,\mathcal{F}, \mathcal{P})$ and let $\mathcal{D}$ be a set of possible CDFs on the sample space $\Omega$. In what follows, we adopt the following definition of a statistical model \citep{McCullagh2002}.
\begin{definition}
A statistical model is a pair $\langle \mathcal{D}, \Omega \rangle$.
\end{definition}

Let $\mathbb{D}$ denote the set of all possible CDFs on $\Omega$. Consider a finite-dimensional parameter set $\Theta$ together with a function $g:\Theta\rightarrow \mathbb{D}$, which assigns to each parameter point $\theta\in\Theta$ a CDF $F_{\theta}$ on $\Omega$. 
\begin{definition}
A parametric statistical model is a triple $\langle \Theta, g, \Omega \rangle$.
\end{definition}
\begin{definition}
A non-parametric statistical model is a pair $\langle \mathbb{D}, \Omega \rangle$.
\end{definition}
Note that semi-parametric models are also possible.

Consider now the outcome $o\in \Omega$ of an experiment. Statistics operates under the assumption that there is a distinct element $d\in \mathcal{D}$ that generates the observed data $o$. The aim of statistical inference is then to determine which element(s) are likely to be the one generating the data. A widely adopted method to carry out statistical inference is hypothesis testing.

In hypothesis testing the statistician selects a significance level $\alpha$ and formulates a null hypothesis, e.g. ``element $d\in\mathcal{D}$ has generated the observed data,'' and an alternative hypothesis, e.g. ``another element in $\mathcal{D}/d$ has generated the observed data.'' Depending on the type of hypothesis formulated, she must then select a suitable statistical test and derive the distribution of the associated test statistic under the null hypothesis, where a statistic is a function $f:\Omega \rightarrow \mathbb{R}$. By using this distribution, one determines the probability $p_o$ of obtaining a test statistic at least as extreme as the one associated with outcome $o$, i.e. the ``$p$-value''. If this probability is less than $\alpha$, this means that the observed result is highly unlikely under the null hypothesis, and the statistician should therefore ``reject the null hypothesis.'' Conversely, if this probability is greater or equal to $\alpha$, the evidence collected is insufficient to support a conclusion against the null hypothesis, hence we say that one ``fails to reject the null hypothesis.'' 

\subsection{Statistical constraints}\label{sec:statistical_constraints}

\begin{definition}
A statistical constraint is a constraint that embeds a parametric or a non-parametric statistical model and a statistical test with significance level $\alpha$ that is used to determine which assignments satisfy the constraint.
\end{definition}

A {\bf parametric statistical constraint} $c$ takes the general form $c(T,g,O,\alpha)$; where $T$ and $O$ are sets of decision variables and $g$ is a function as defined in Section \ref{sec:inference}. Let $T\equiv \{t_1,\ldots,t_{|T|}\}$, then $\Theta=D(t_1)\times\ldots\times D(t_{|T|})$. Furthermore, let $O\equiv\{o_1,\ldots,o_{|O|}\}$, then $\Omega= D(o_1)\times\ldots\times D(o_{|O|})$. An assignment is consistent with respect to $c$ if the statistical test fails to reject the associated null hypothesis, e.g. ``$F_\theta$ generated $o_1,\ldots,o_{|O|}$,'' at significance level $\alpha$.

A {\bf non-parametric statistical constraint} $c$ takes the following general form $c(O_1,\ldots,O_k,\alpha)$; where $O_1,\ldots,O_k$ are sets of decision variables. Let $O_i\equiv\{o^i_1,\ldots,o^i_{|O_i|}\}$, then $\Omega=\bigcup_{i=1}^k D(o^i_1)\times\ldots\times D(o^i_{|O_i|})$. An assignment is consistent with respect to $c$ if the statistical test fails to reject the associated null hypothesis, e.g. ``$\{o^1_1,\ldots,o^1_{|O_1|}\}$,\ldots,$\{o^k_1,\ldots,o^k_{|O_k|}\}$ are drawn from the same distribution,'' at significance level $\alpha$.

In contrast to classical statistical testing, {\em random variates}, i.e. random variable realisations $(\omega'_1,\ldots,\omega'_n)^{T}$, associated with a sample are modelled as decision variables. The {\em sample}, i.e. the set of random variables $(\omega_1,\ldots,\omega_n)^{T}$ that generated the random variates is not explicitly modelled. 

\subsubsection{Student's $t$ test constraint}\label{sec:student}

Recall that a $t$-test is any statistical hypothesis test in which the test statistic follows a Student's $t$ distribution \citep{Student1908} if the null hypothesis is supported. The classic one-sample $t$-test compares the mean of a sample to a specified mean. We consider the null hypothesis $H_0$ that ``the sample is drawn from a random variable with mean $\mu$.'' 
The test statistic is 
\[t=\frac{m-\mu}{s/\sqrt{n}}\]
where $m$ is the sample mean, $s$ is the sample standard deviation and $n$ is the sample size. Since Student's $t$ distribution is symmetric, $H_0$ is rejected if $\Pr(x>t|H_0)<\alpha/2$ or $\Pr(x<t|H_0)<\alpha/2$ that is
\[\mu< m+\frac{s}{\sqrt{n}} T^{-1}_{n-1}(\alpha/2)~~~~\text{or}~~~~\mu> m-\frac{s}{\sqrt{n}} T^{-1}_{n-1}(\alpha/2)\]
where $T^{-1}_{n-1}$ is the inverse Student's $t$ distribution with $n-1$ degrees of freedom. The respective single-tailed tests can be used to determine if the sample is drawn from a random variable with mean less (greater) than $\mu$.

The Student's $t$ test constraint \citep{rossi2014} is defined as
\[t\text{-test}^\alpha_{w}(x,\mu)\]
where $x_i$, for $i=1\ldots,n$ is a set of decision variables each of which represents a random variate $\omega'_i$; $\mu$ is a decision variable representing the mean of the random variable $\omega$ that generated the sample. Parameter $\alpha\in(0,1)$ is the significance level; parameter $w\in\{\leq,\geq,=,\neq\}$ identifies the type of statistical test that should be employed, e.g. ``$\leq$'' refers to a single-tailed Student's $t$-test that determines if the mean of $\omega$ is less than or equal to $\mu$,``$=$'' refers to a two-tailed Student's $t$-test that determines if the mean of $\omega$ is equal to $\mu$, etc. An assignment $\bar{o}_1,\ldots,\bar{o}_n,\bar{\mu}$ satisfies $t\text{-test}^\alpha_{w}$ if and only if a one-sample Student's $t$-test fails to reject the null hypothesis identified by $w$; e.g. if $w$ is ``$=$'', then the null hypothesis is `` the mean of the random variable that generated $\bar{o}_1,\ldots,\bar{o}_n$ is equal to $\bar{\mu}$.''

In \citep{rossi2014}, the authors also introduced the two-sample $t$-test, which compares means $\mu_1$ and $\mu_2$ of two samples; but they did not discuss filtering strategies for the Student's $t$ test constraint. However, this constraint can be easily decomposed by employing a \textsc{mean} constraint and a \textsc{standard\_error} constraint as shown in Fig. \ref{model:ttest_decomposition}.

\begin{figure}[h!]
\begin{center}
        \framebox{
        \begin{tabular}{ll}
        \mbox{\bf Constraints:}\\
        ~~~~(1)~~$\mu \leq m + s T^{-1}_{n-1}(\alpha/2)$\\
        ~~~~(2)~~$\mu \geq m - s T^{-1}_{n-1}(\alpha/2)$\\        
        \multicolumn{2}{l}{
        ~~~~(2)~~$\textsc{mean}(O;m)$}\\
        \multicolumn{2}{l}{
        ~~~~(2)~~$\textsc{standard\_error}(O;s)$}\\
        \\     
        \mbox{\bf Parameters:} \\
        ~~~~$T^{-1}$ 						&inverse $t$ distribution\\
        ~~~~$n$							&number of random variates\\                        							                   
        \\
        \mbox{\bf Decision variables:} \\
        ~~~~$O$							&random variates\\        
        ~~~~$\mu$						&``true mean''\\
        ~~~~$m$							&sample mean\\
        ~~~~$s$							&standard error
        \end{tabular}
        }
    \caption{t\text{-test}${}^{\alpha}_{=}$ decomposition}
    \label{model:ttest_decomposition}
\end{center}    
\end{figure}

The Student's $t$ test constraint is an example of parametric statistical constraint. An example of a non-parametric statistical constraint is the non-parametric Kolmogorov-Smirnov constraint, which is also discussed in \citep{rossi2014}.

\section{Relevant counting and matrix constraints}\label{sec:counting_matrix}

In Constraint Programming (CP) \citep{1207782} counting constraints and value constraints represent two important constraint classes which can be used to model a wide range of practical problems. In this section, we introduce two counting constraints, the {\normalfont\scshape bin\_counts} and the {\normalfont\scshape contingency} constraints that are relevant in the context of our discussion. We also introduce the {\normalfont\scshape matrix\_inversion} constraint, that is relevant in the context of multivariate declarative statistics.

\subsection{The {\normalfont\scshape bin\_counts} constraint}\label{sec:bincount}

Given a list of numbers, counting the number of elements whose values lie in successive bins of given widths represents a problem often faced in practice.  

Consider a list of $n$ values and a set of $m$ bins covering half-open intervals $[b_j,b_{j+1})$, $j=1,\ldots,m$. Our aim is to count, for each bin, the number of elements in the list whose values lie in it. 

{\bf Example.} Consider the following list of $n=10$ values $\{1, 1, 5, 3, 1, 2, 1, 1, 3, 1\}$ and a set of $m=3$ bins covering intervals $[1,3)$, $[3,4)$, $[4,6)$. By using command {\verb BinCounts } in Mathematica\textsuperscript{TM} we obtain the following counts for the 3 bins considered: $\{7,2,1\}$.

Note that, in the most general case, both values and bin sizes can be real values. Let $b_1,\ldots,b_{m+1}$ be scalar values; $x_i$ for $i=1,\ldots,n$ be decision variables with domain $\text{Dom}(x_i)$; and $c_j$ for $j=1,\ldots,m$ be decision variables with domain $\text{Dom}(c_j)$.

\begin{definition}
$\textsc{bin\_counts}_b(x;c)$ holds iff $c_j$ is equal to the count of values assigned to $x_1,\ldots,x_n$ which lie within interval $[b_j,b_{j+1})$.
\end{definition} 

In what follows we discuss a possible decomposition for this constraint. The $\textsc{global\_cardinality}_v$($x;c$) constraint \citep{charme89} requires that, for each $j=1,\ldots,m$, decision variable $c_j$ is equal to the number of variables $x_1,\ldots,x_n$ that are assigned scalar $v_j$. As shown in Fig. \ref{model:bincounts_decomposition}, we decompose \textsc{bin\_counts} by means of $n$ auxiliary variables $a_i$ such that $\text{Dom}(a_i)\in\{1,\ldots,m\}$ and a \textsc{global\_cardinality} constraint. Essentially, we map each variable $x_i$ to its bin $a_i$ (constraint 1), and then we count occurrences $c_j$ by exploiting the \textsc{global\_cardinality} constraint (constraint 2). Note that $x_i$ may take values that fall outside the range of values covered by bins. 

\begin{figure}[h!]
\begin{center}
        \framebox{
        \begin{tabular}{ll}
        \mbox{\bf Constraints:}\\
        ~~~~(1)~~$a_i = j \leftrightarrow x_i \geq b_j \wedge x_i < b_{j+1}$	&$i=1,\ldots,n$\\
        \multicolumn{2}{l}{
        ~~~~(2)~~$\textsc{global\_cardinality}_{1,\ldots,m}(a;c)$}\\
        \\     
        \mbox{\bf Parameters:} \\
        ~~~~$b_1,\ldots,b_{m+1}$ 			&bin boundaries\\
        ~~~~$n$							&number of value variables\\                        							
        \\
        \mbox{\bf Decision variables:} \\
        ~~~~$x_i$							&value variables\\        
        ~~~~$a_i$							&value-bin allocation\\        
        ~~~~$c_j$ 						&bin counts
        \end{tabular}
        }
    \caption{\textsc{bin\_counts} decomposition}
    \label{model:bincounts_decomposition}
\end{center}    
\end{figure}

The reader should note that other decompositions are possible. One may, for instance, employ a hierarchical GCC constraint \citep{citeulike:14398655} and model the problem as a problem of counting on a tree of depth two; or develop a decomposition based on domain views to compute the bin \citep{citeulike:14398654}. Furthermore, a related work in which dedicated filtering strategies are discussed for a similar constraint, i.e. \textsc{frequency}, is \citep{citeulike:14393881}. However, investigating the relative performance of these different modeling strategies falls beyond the scope of our work; we therefore leave this investigation as a future research direction.

\subsection{The {\normalfont\scshape contingency} constraint}\label{sec:contingency}

Contingency tables \citep{pearson1906theory} are often used in statistics to carry out tests such as the well-known $\chi^2$ test of independence \citep{doi:10.1080/14786440009463897}.

Consider a list of $n$ paired values $\langle v^k_1, v^k_2 \rangle$, for $k=1,\ldots,n$ and two sets of bins covering half-open intervals $[b_i,b_{i+1})$, for $i=1,\ldots,m_1$ and $[d_j,d_{j+1})$, for $j=1,\ldots,m_2$. A contingency table counts, for every possible pair $\langle[b_i,b_{i+1}), [d_j,d_{j+1})\rangle$, the number $c_{ij}$ of elements $\langle v^k_1, v^k_2 \rangle$ such that $v^k_1\in [b_i,b_{i+1})$ and $v^k_2\in [d_j,d_{j+1})$. Moreover, it also counts the marginal values $h_i = \sum_{j=1}^{m_2} c_{ij}$, for all $i=1,\ldots,m_1$; and $w_j = \sum_{i=1}^{m_1} c_{ij}$, for all $j=1,\ldots,m_2$.

{\bf Example.} Consider the following list of $n=3$ values $\{\langle 1, 2 \rangle, \langle 3, 1 \rangle, \langle 3, 2 \rangle\}$ and two sets of $m=2$ bins covering intervals $[1,3)$, $[3,4)$ and $[1,2)$, $[2,4)$, respectively. The associated contingency table is shown in Table \ref{tab:contingency}.
\begin{table}[h!]
\center
\begin{tabular}{c|cc|c}
		&$[1,2)$	&$[2,4)$	&$h_i$\\
\hline
$[1,3)$	&0		&1		&1\\
$[3,4)$	&1		&1		&2\\
\hline
$w_j$		&1		&2
\end{tabular}
\caption{A contingency table}
\label{tab:contingency}
\end{table}

Note that, in the most general case, both values and bin sizes can be real values.

Let $b_1,\ldots,b_{m_1+1}$ and $d_1,\ldots,d_{m_2+1}$ be scalar values; $\langle x^k_1, x^k_2\rangle$ for $k=1,\ldots,n$ be pairs of decision variables with domain $\text{Dom}(x^k_1),\text{Dom}(x^k_2)$; $c_{ij}$ for $i=1,\ldots,m_1$ and $j=1,\ldots,m_2$ be decision variables with domain $\text{Dom}(c_{ij})$;  $h_i$ for $i=1,\ldots,m_1$ be decision variables with domain $\text{Dom}(h_i)$, and $w_j$ for $j=1,\ldots,m_2$ be decision variables with domain $\text{Dom}(w_j)$.

\begin{definition}
$\textsc{contingency}_{b,d}(\langle x_1, x_2\rangle;c;h;w)$ holds iff $c_{ij}$ is equal to the number of occurrences in $\langle x^k_1, x^k_2\rangle$ for $k=1,\ldots,n$ such that $x^k_1\in [b_i,b_{i+1})$ and $x^k_2 \in [d_j,d_{j+1})$; $h_i = \sum_{j=1}^{m_2} c_{ij}$, for all $i=1,\ldots,m_1$; and $w_j = \sum_{i=1}^{m_1} c_{ij}$ for all $j=1,\ldots,m_2$.
\end{definition} 

In what follows we discuss a possible decompositions for this constraint. As shown in Fig. \ref{model:contingency_decomposition}, we decompose \textsc{contingency} by reification of a condition checking if a given pair of values $\langle x^k_1, x^k_2\rangle$ for $k=1,\ldots,n$ belongs to $\langle[b_i,b_{i+1}), [d_j,d_{j+1})\rangle$ for $i=1,\ldots,m_1$ and $j=1,\ldots,m_2$ (constraint 1). Constraints 2 and 3 compute marginals $h_i$ and $w_j$. Also in this case, we may admit pairs $\langle x^k_1, x^k_2\rangle$ taking values that fall outside the range of values covered by bins. 

\begin{figure}[h!]
\begin{center}
        \framebox{
        \begin{tabular}{ll}
        \mbox{\bf Constraints:}\\
        ~~~~(1)~~$c_{ij} = \sum_{k=1}^n (x^k_1\in [b_i,b_{i+1}) \wedge x^k_2\in [d_j,d_{j+1}))$	&$i=1,\ldots,m_1$;\\
        																		&$j=1,\ldots,m_2$\\        
        ~~~~(2)~~$h_i = \sum_{j=1}^{m_2} c_{ij}$	&$i=1,\ldots,m_1$\\
        ~~~~(3)~~$w_j = \sum_{i=1}^{m_1} c_{ij}$	&$j=1,\ldots,m_2$\\        
        \\     
        \mbox{\bf Parameters:} \\
        ~~~~$n$									&value list length\\         
        ~~~~$b_1,\ldots,b_{m_1+1},d_1,\ldots,d_{m_2+1}$ 	&bin boundaries\\						                         
        \\
        \mbox{\bf Decision variables:} \\
        ~~~~$\langle x^k_1, x^k_2\rangle$		&value variables\\        
        ~~~~$c_{ij}$ 						&bin counts
        \end{tabular}
        }
    \caption{\textsc{contingency} decomposition}
    \label{model:contingency_decomposition}
\end{center}    
\end{figure}

\subsection{The {\normalfont\scshape matrix\_inversion} constraint}\label{sec:matrixinversion}

Matrix inversion \citep[see e.g.][]{citeulike:14405083} is an essential tool in multivariate statistical analysis \citep{citeulike:14214028}. The inverse of a square matrix $\boldsymbol{A}$, sometimes called a reciprocal matrix, is a matrix $\boldsymbol{A}^{-1}$ such that $\boldsymbol{A}\boldsymbol{A}^{-1}=\boldsymbol{I}$, where $\boldsymbol{I}$ is the identity matrix. A square matrix admits an inverse iff its determinant is nonzero.

To construct a {\normalfont\scshape matrix\_inversion} constraint we apply symbolic Gauss-Jordan elimination to a matrix of the form
\[
[\boldsymbol{A}~\boldsymbol{I}]\equiv
\left[
\begin{array}{ccccccc}
a_{11}&\ldots&a_{1n}&1&0&\ldots&0\\
a_{21}&\ldots&a_{2n}&0&1&\ldots&0\\
\vdots&\ddots&\vdots&\vdots&\vdots&\ddots&\vdots\\
a_{n1}&\ldots&a_{nn}&0&0&\ldots&1\\
\end{array}\right]
\]
in order to obtain algebraic expressions for elements $b_{11},\ldots,b_{nn}$ belonging to the resulting matrix of the form
\[
\left[
\begin{array}{ccccccc}
1&0&\ldots&0&b_{11}&\ldots&b_{1n}\\
0&1&\ldots&0&b_{21}&\ldots&b_{2n}\\
\vdots&\ddots&\vdots&\vdots&\vdots&\ddots&\vdots\\
0&0&\ldots&1&b_{n1}&\ldots&b_{nn}\\
\end{array}\right]
\equiv[\boldsymbol{I}~\boldsymbol{A}^{-1}]
\]
We post these expressions in Choco by relying on the interval arithmetic extensions discussed in \citep{tricks2013}, which makes it possible to model real variables and constraints, and delegate associated reasoning during search to Ibex. 

\begin{definition}
Let $\boldsymbol{A}$ be a square $n\times n$ matrix of real valued decision variables; {\normalfont\scshape matrix\_inversion}$(\boldsymbol{A}, \boldsymbol{A}^{-1})$ holds iff $\boldsymbol{A}\boldsymbol{A}^{-1}=\boldsymbol{I}$.
\end{definition} 

Note that, if $\boldsymbol{A}$ is singular and does not admit an inverse, the associated  {\normalfont\scshape matrix\_inversion} constraint leads to infeasibility.

{\bf Example.} Consider a $2\times 2$ matrix $\boldsymbol{A}$ with elements $a_{11},a_{12},a_{21},a_{31}$, by applying symbolic Gauss-Jordan elimination we obtain the inverse matrix $\boldsymbol{A}^{-1}$ elements $b_{11}=a_{22}/(-a_{12} a_{21} + a_{11} a_{22})$; $b_{12}=-a_{12}/(-a_{12} a_{21} + a_{11} a_{22})$; $b_{12}=-a_{21}/(-a_{12} a_{21} + a_{11} a_{22})$; and $b_{22}=a_{11}/(-a_{12} a_{21} + a_{11} a_{22})$.

\section{A selection of statistical constraints}\label{sec:selection_statistical_constraints}

In this section we introduce a range of new statistical constraints that, as we will discuss in Section \ref{sec:applications}, represent effective modeling tools in the context of declarative statistics. We first summarize the key elements of our toolbox.

The $t$-test statistical constraints, originally introduced in \citep{rossi2014} and discussed in Section \ref{sec:student}, aims at capturing Student's $t$ statistic, which can be used to compare the mean of a sample against a target value (parametric), or against the mean of a second sample (non-parametric).

The $\chi^2$ goodness of fit statistical constraint (Section \ref{sec:chi_squared}) can be used to compare the distribution of a sample against a parametric reference distribution. Another statistical constraint that can be used for a similar purpose when the target distribution is continuous is the Kolmogorov-Smirnov statistical constraints, originally introduced in \citep{rossi2014}.

The $\chi^2$ test of independence statistical constraint (Section \ref{sec:chi_squared_ind}) can be used to assess independence between two samples.

Fisher's ratio statistical constraint (Section \ref{sec:fisher}) can be used to compare the variance of a sample against a reference value, or the variances of two samples.

Hotelling's statistical constraints (Section \ref{sec:hotelling_t}) can be used to compare the mean of a multivariate sample against a target mean vector when the sample variance-covariance matrix is known (Hotelling $\chi^2$) or unknown (Hotelling $t^2$). 

\subsection{The $\chi^2$ goodness of fit statistical constraint}\label{sec:chi_squared}

Recall that the $\chi^2$ distribution with $k$ degrees of freedom is the distribution of a sum of the squares of k independent standard normal random variables.

The $\chi^2$ constraint captures Pearson's $\chi^2$ statistic, which can be employed, for instance, to carry out a goodness of fit test \citep{doi:10.1080/14786440009463897} to establish whether an observed frequency distribution differs from a theoretical distribution.

Let $x_i$, for $i=1,\ldots,n$, be a decision variable that represents a random variate; $c_j$, $j=1,\ldots,m$, be a decision variable that represents the number of variables $x_i$ which take a value in $[b_j,b_{j+1})$; $t_j$ be a decision variable that represents the theoretical reference count for bin $[b_j,b_{j+1})$ --- note that reference counts may be an empirical distribution or a function of the distribution of the random variable that generated the sample; and finally, let $s$ be the $\chi^2$ statistic. 

\begin{definition}
$\chi^2_b(x;t;s)$ holds iff $\textsc{bin\_counts}_b(x;c)$ and $s=\sum_{i=1}^m (c_i - t_i)^2/t_i$.
\end{definition}

It is worth observing that a small $\chi^2$ statistic reflects a small difference between the theoretical reference counts and the observed counts.

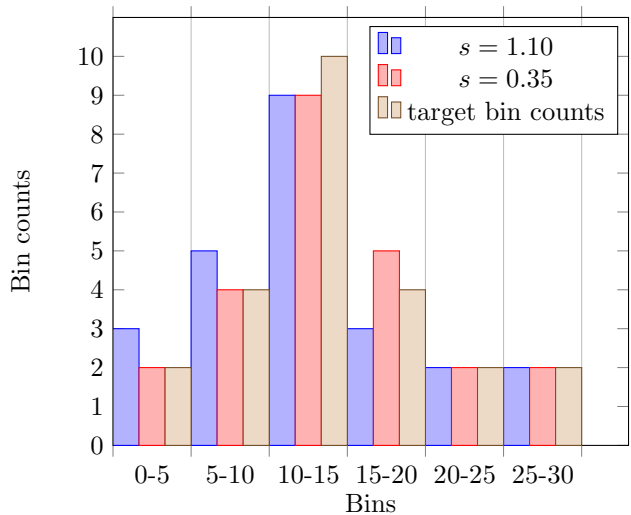
\begin{figure}
\centering
\begin{tikzpicture}
\begin{axis}[
    ytick={0,1,2,3,4,5,6,7,8,9,10},
    xlabel=Bins,
    ylabel=Bin counts,
    ybar interval,
    ymin=0,
    xmin=0,
    xticklabel=\pgfmathprintnumber\tick-\pgfmathprintnumber\nexttick
]
\addplot +[
    hist={
        bins=6,
        data min=0,
        data max=30
    }   
]
table[y=value] {
value
13
7
6
12
10
14
8
19
5
3
20
12
18
11
11
17
12
2
12
28
6
25
21
1
};
\addlegendentry{$s= 1.10$}
\addplot +[
    hist={
        bins=6,
        data min=0,
        data max=30
    }   
]
table[y=value] {
value
26
9
6
12
16
10
8
28
5
21
0
12
1
12
11
10
17
19
24
11
16
19
13
11
};
\addlegendentry{$s= 0.35$}
\addplot coordinates {(0, 2) (5, 4) (10, 10) (15, 4) (20, 2) (25, 2) (30, 2)};
\addlegendentry{target bin counts}
\end{axis}
\end{tikzpicture}
\caption{Sample bincounts obtained in the context of our numerical example for different values of $\chi^2$. Note that a smaller $\chi^2$ statistic reflects bin counts that are closer to target counts for each bin, since the $\chi^2$ statistic is essentially an Euclidean distance in $m$ dimensions.}
\label{fig:bincounts_chi}
\end{figure}

{\bf Example.} We consider a problem with $n=24$ variables $x_i$ and $m=6$ bins such that $b_i=5(i-1)$ for $i=1,\ldots,m+1$. $\text{Dom}(x_i)$ comprises values $\{0,\ldots,30\}$. $\text{Dom}(c_j)$ comprises values $\{0,\ldots,n\}$. We implemented the model and set as a goal the instantiation of variables $x_1,\ldots,x_n$. Theoretical reference counts $t_j$ for the $m$ bins are $t=\{2,4,10,4,2,2\}$. The bin counts for two solutions are shown in Fig. \ref{fig:bincounts_chi} and contrasted against theoretical reference counts. 

If relevant assumptions associated with the $\chi^2$ goodness of fit test are met, this test can be modelled by forcing the statistic $s$ to remain below the $1-\alpha$ quantile $Q$ of a $\chi^2$ distribution with $m-1$ degrees of freedom, where $\alpha$ is the significance level for the test; this leads to the $\chi^2$ statistical constraint, which is defined as follows.
\begin{definition}
$\chi^2_{b,\alpha}(v;t)$ holds iff $\chi^2_b(v;t;s)$ and $s\leq Q$.
\end{definition}
Corrections such as the one proposed by \cite{10.2307/2983604} for dealing with low or zero observed bincounts can be easily embedded in our decomposition.

\subsection{The $\chi^2$ test of independence statistical constraint}\label{sec:chi_squared_ind}

In addition to a goodness of fit test, the $\chi^2$ statistic can be employed to carry out a test of independence. 

Consider a list of $n$ paired decision variables $\langle x^k_1, x^k_2 \rangle$, for $k=1,\ldots,n$, representing paired random variates; two sets of bins covering half-open intervals $[b_i,b_{i+1})$, for $i=1,\ldots,m_1$ and $[d_j,d_{j+1})$, for $j=1,\ldots,m_2$; and let $s$ be the $\chi^2$ statistic. 

\begin{definition}
$\chi^2_{b,d}(\langle x_1,x_2\rangle;s)$ holds iff $\textsc{contingency}_{b,d}(\langle x_1, x_2\rangle;c;h;w)$ and 
\[s=\sum_{i=1}^n \sum_{j=1}^n \frac{(c_{ij} - E_{ij})^2}{E_{ij}}\]
where $E_{ij}=h_i w_j/n$.
\end{definition}

If relevant assumptions associated with the $\chi^2$ test of independence are met, this test can be modelled by forcing the statistic $s$ to remain below the $1-\alpha$ quantile $Q$ of a $\chi^2$ distribution with $(m_1-1)(m_2-1)$ degrees of freedom, where $\alpha$ is the significance level for the test; this leads to the $\chi^2$ statistical constraint, which is defined as follows.
\begin{definition}
$\chi^2_{b,d,\alpha}(\langle x_1,x_2\rangle)$ holds iff $\chi^2_{b,d}(\langle x_1,x_2\rangle;s)$ and $s\leq Q$.
\end{definition}
Once more, corrections for dealing with low or zero observed bincounts can be easily embedded in our decomposition.

\subsection{Fisher's ratio statistical constraint}\label{sec:fisher}

Recall that a random variable distributed according to an $F$-distribution with parameters $k_1$ and $k_2$ arises as the ratio of two independent random variables following chi-squared distributions with degrees of freedom $k_1$ (numerator) and $k_2$ (denominator).

Fisher's ratio statistic \citep[see e.g.][]{citeulike:14403808} is widely used to compare variances of two normally distributed populations under the null hypothesis that variances are equal. This statistic can be used in the context of a two-tailed test or a one-tailed test. The former tests against the alternative hypothesis that the variances are not equal; the latter may be used to tests against the alternative hypothesis that the variance from the first population is either greater than or less than (but not both) the second population variance.

Consider two lists of $n_1$ and $n_2$ decision variables $x^i_1$ and $x^j_2$, where $i=1\ldots,n_1$ and $j=1,\ldots,n_2$; and let $s$ be the $F$ statistic.
\begin{definition}
$F(x_1;x_2;s)$ holds iff $\textsc{variance}(x_1;v_1)$, $\textsc{variance}(x_2;v_2)$ and 
\[s=v_1/v_2.\]
\end{definition}

If relevant assumptions associated with the $F$ test are met; this test can be modelled by forcing the statistic $s$ to remain 
\begin{itemize}
\item {\bf two-tailed test} (alternative hypothesis: $v_1\neq v_2$). Between the $\alpha/2$ quantile $Q_{lb}$ and the $1-\alpha/2$ quantile $Q_{ub}$ of an $F$ distribution \citep{snedecor1934calculation} with $n_1-1$ (numerator) and $n_2-1$ (denominator) degrees of freedom;
\item {\bf one-tailed test} (alternative hypothesis: $v_1>v_2$). Below the $1-\alpha$ quantile $Q$ of an $F$ distribution with $n_1-1$ and $n_2-1$ degrees of freedom; 
\item {\bf one-tailed test} (alternative hypothesis: $v_1<v_2$). Above the $\alpha$ quantile $Q$ of an $F$ distribution with $n_1-1$ and $n_2-1$ degrees of freedom;
\end{itemize}
where $\alpha$ is the significance level for the test. This leads to the $F$ statistical constraint, which is defined as follows.
\begin{definition}
$F_\alpha(x_1;x_2)$ holds iff $F(x_1;x_2;s)$ and the statistic satisfies the relevant one-tailed or two-tailed test.
\end{definition}

\subsection{Hotelling's $t^2$ statistical constraint}\label{sec:hotelling_t}

Hotelling's $T^2$ distribution is a multivariate generalisation of the $\chi^2$ distribution. Hotelling's $t^2$ statistic is used in multivariate hypothesis testing to assess differences between mean vectors of different samples \citep{citeulike:10382072}.

Consider a list $\langle X\rangle$ of $n$ tuples $\langle x^k_1, \ldots, x^k_p \rangle$, for $k=1,\ldots,n$ --- where $x^k_1, \ldots, x^k_p$ are decision variables --- representing independent random variates drawn from a $p$-dimensional random vector $\omega$. Let $\mu$ be a vector of decision variables $\mu_1,\ldots,\mu_p$ representing the mean of $\omega$ and $\Sigma$ --- a $p\times p$ matrix of decision variables --- be the variance-covariance matrix of $\omega$. We define a constraint capturing Hotelling's $\chi^2$ statistic.

\begin{definition}
$\chi^2(\langle X\rangle;\mu;\Sigma;s)$ holds iff $s=n(\bar{x}-\mu)'\Sigma^{-1}(\bar{x}-\mu)$; where $\bar{x}$ is a vector with elements $\bar{x}_1,\ldots,\bar{x}_p$, such that \textsc{mean}$(x_i,\bar{x}_i)$ and $x_i$ is a list with elements $x^1_i,\ldots,x^n_i$.
\end{definition}

Note that this constraint is easily decomposed into an algebraic expression once $\Sigma^{-1}$ has been obtained by posting a {\normalfont\scshape matrix\_inversion} constraint on $\Sigma$.

If relevant assumptions are met --- i.e. $n$ is large and/or $\omega$ is normally distributed --- Hotelling's $\chi^2$ test can be modelled by forcing the statistic $s$ to remain below the $1-\alpha$ quantile of a $\chi^2$ distribution with $p$ degrees of freedom; thus leading to the $\chi^2_\alpha(\langle X\rangle;\mu;\Sigma;s)$ statistical constraint.

In some cases, the variance-covariance matrix $\Sigma$ of $\omega$ must be estimated from the data. We then proceed by replacing $\Sigma$ with the sample variance-covariance matrix $S$ and by defining as follows a constraint capturing Hotelling's $t^2$ statistic.

\begin{definition}
$t^2(\langle X\rangle;\mu;s)$ holds iff $s=n(\bar{x}-\mu)'S^{-1}(\bar{x}-\mu)$, where $S$ is a matrix of decision variables $S_{ij}$ such that \textsc{covariance}$(x_i,x_j,S_{ij})$ for $i,j=1,\ldots,p$.
\end{definition}

Once more, this constraint is easily decomposed into an algebraic expression once $S^{-1}$ has been obtained by posting a {\normalfont\scshape matrix\_inversion} constraint on $S$.



If relevant assumptions are met, Hotelling's $t^2$ test can be modelled by forcing the statistic $s$ to remain below the $1-\alpha$ quantile $Q$ of Hotelling's $T^2_{p,n-1}$ distribution, where $p$ is the dimensionality parameter and $m$ are the degrees of freedom; thus leading to the $t^2_\alpha(\langle X\rangle;\mu;s)$ statistical constraint.

\begin{definition}
$t^2_\alpha(\langle X\rangle;\mu)$ holds iff $t^2(\langle X\rangle;\mu;s)$ and $s\leq Q$.
\end{definition}

Note that, like in the case of the $t$-test statistical constraint, this constraint is also easily extended if one aims to compare the means of two samples.

\section{Applications of declarative statistics}\label{sec:applications}

In this section we showcase applications for the statistical constraints we previously introduced. The section is structured as shown in Table \ref{tab:applications}.

\begin{table}[h!]
\begin{tabular}{lll}
{\em Section}		&{\em Application}		&{\em Statistical constraint(s)}\\
\hline
\ref{sec:regression}	&Linear model fitting		&$\chi^2$  goodness of fit\\
\hline
\ref{sec:time_series}	&Time series analysis	&$\chi^2$  goodness of fit\\
									&&$\chi^2$  test of independence\\
\hline									
\ref{sec:anova}		&ANOVA								&Fisher's ratio\\
				&Comparing means of two or more samples	&Hotelling's $t^2$\\
\hline				
\ref{sec:multinomial}	&multinomial proportions conf. interv.	&Hotelling's $\chi^2$ and $t^2$	\\				
\hline
\end{tabular}
\caption{Applications considered and associated statistical constraints.}
\label{tab:applications}
\end{table}

\subsection{Linear model fitting}\label{sec:regression}



To motivate declarative statistics, we employ a simple example from classical statistic: linear model fitting and associated statistical analysis \citep[see e.g.][]{citeulike:14403808}.

Consider a set of $T$ random variates $v_t$ for $t=1,\ldots,T$. For simplicity, assume that variates have been generated according to 
\[v_t=a t+b+e_t\]
where $e_t$ is a random variable with cumulative distribution function $F_\sigma$. Without loss of generality and in line with established practices in statistics, we shall assume that $F_\sigma$ is normal with mean zero and standard deviation $\sigma$.

\begin{figure}[h!]
\centering
\includegraphics{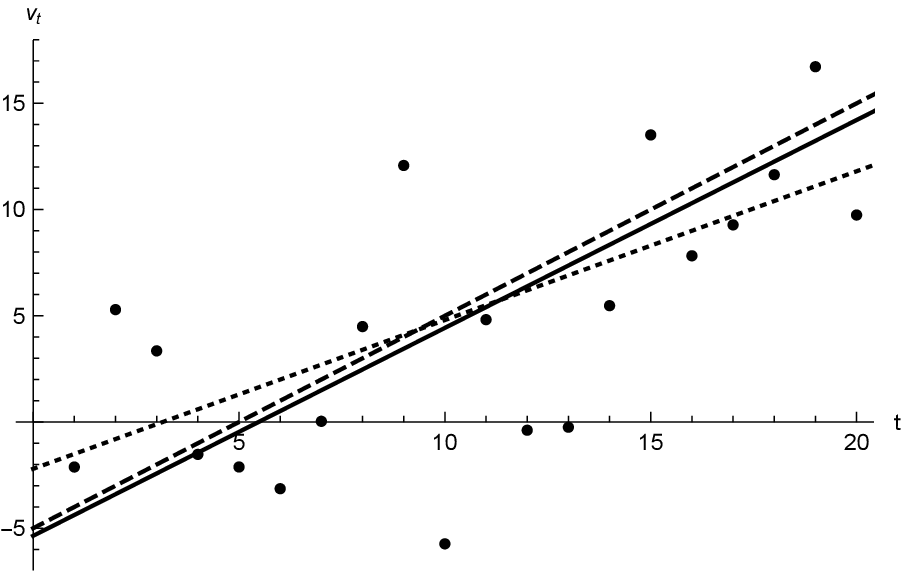}
\caption{Linear model fitting. The dots represent random variates in Table \ref{tab:example_1_random_variates}; the solid line is the model that generated these variates, which we are trying to estimate. The dotted line, is the model obtained via the method of least squares. The dashed line is the model obtained via declarative statistics, by solving the model in Fig. \ref{model:example_1_linear_regression_dscp_model}.}
\label{fig:example_1_linear_regression}
\end{figure}

To focus attention, we will refer to the variates $(t,v_t)$ in Table \ref{tab:example_1_random_variates}, which have been generated by setting model parameters as follows: $a=1$, $b=-5$, $\sigma=5$, and $T=20$ (Fig. \ref{fig:example_1_linear_regression} - solid line). By fitting a linear model $y=ax+c$ via the method of least squares we obtain parameters $a=0.700$ and $b=-2.203$ (Fig. \ref{fig:example_1_linear_regression} - dotted line).

\begin{table}[h!]
\centering
\begin{tabular}{lll}
$(1,-2.119453760526702)$&$ (2,5.290814814602713)$&$ (3,3.3477370212059263)$\\
$(4,-1.524427844666869$&$ (5,-2.11767611724241$&$ (6,-3.1393019984876567)$\\
$(7,0.031583398832589316)$&$ (8,4.492170566086558)$&$ (9,12.075689120209544)$\\
$(10,-5.734583134742884)$&$ (11,4.817685166491335)$&$ (12,-0.38732268295202754)$\\
$(13,-0.2451087678267534)$&$ (14,5.476406521028064)$&$ (15,13.513668326933141)$ \\
$(16,7.824341452223766)$&$ (17,9.279650356164751)$&$ (18,11.640247250501195)$ \\
$(19,16.724560475349527)$&$ (20,9.74407257497221)$
\end{tabular}
\caption{Random variates for our numerical example}
\label{tab:example_1_random_variates}
\end{table}

We next apply declarative statistics to this simple example. We set up a constraint programming model as shown in Fig. \ref{model:example_1_linear_regression_dscp_model}. The model takes as input the random variates, which are given, a list of bin bounds and target counts for each bin, which represent the target distribution for the errors.\footnote{Note that these are errors, not residuals, since our model does not rely on least square estimates of parameters.} Constraint (1) determines the value of the errors ($e_t$) as a linear function of the random variates ($v_t$), slope ($a$) and intercept ($b$); constraint (2) computes the target counts for each bin as a function of the standard deviation $\sigma$ of a normal cumulative distribution with zero mean;\footnote{Note that to implement this in Ibex we relied on analytical approximations of the normal cumulative distribution function, see \citep[][eq. 4.5]{citeulike:14402464}.} constraint (3) computes the $\chi^2$  goodness of fit statistic $s$, which is then minimised in the objective function; constraint (4) forces the chi squared statistic $s$ to remain below the $1-\alpha$ quantile of the inverse $\chi^2$ distribution with $m-1$ degrees of freedom.

In the context of this example we set $\alpha=0.05$ and employ 5 bins of size 4, spanning from value -10 to 10: the rationale is to cover value 0 and to allow sufficient scope for error variability; since the statistical test we employ is a chi squared goodness of fit, the bin structure adopted is necessarily arbitrary.\footnote{To ensure that the model remains feasible even when some errors fall outside the given bin structure, one can adopt a {\normalfont\scshape bin\_counts} decomposition that allows the counts to add up to a total that is less or equal to the number of observations; implementations of \textsc{global\_cardinality} that allow this are available in constraint solvers.} The solution obtained by solving this model is $a=0.979$, $b=-5.36$, and $\sigma=4.71$ (Fig. \ref{fig:example_1_linear_regression} - dashed line). 

\begin{figure}[h!]
\begin{center}
        \framebox{
        \begin{tabular}{ll}
        \mbox{\bf Objective:}\\
        ~~~~$\min$ $s$\\
        \\
        \mbox{\bf Constraints:}\\
	~~~~(1)~~$e_t=v_t-(a t+b)$			&for $t=1,\ldots,T$\\
	~~~~(2)~~$\tau_i=T (F_\sigma(b_{i+1})-F_\sigma(b_i))$	&for $i=1,\ldots,m$\\
        \multicolumn{2}{l}{
        ~~~~(3)~~$\chi^2_b(e;\tau;s)$}\\
        ~~~~(4)~~$s\leq Q$\\        
        \\     
        \mbox{\bf Parameters:} \\
        ~~~~$T$							&time periods\\
        ~~~~$v_1,\ldots,v_T$				&random variates\\
        ~~~~$m$							&number of bins\\
        ~~~~$b_1,\ldots,b_{m+1}$ 			&bin boundaries\\
        ~~~~$F_t$ 						&normal cumulative distribution \\ 
        					   				&with mean 0 and standard deviation $\sigma$ \\
	~~~~$Q$							&$1-\alpha$ quantile of the inverse $\chi^2$\\
        									&distribution with $m-1$ degrees\\ 
	        					   			&of freedom\\									
        \\
        \mbox{\bf Decision variables:} \\
        ~~~~$a$							&fitted model slope\\
        ~~~~$b$							&fitted model intercept\\        
        ~~~~$\sigma$						&normal standard deviation\\
        ~~~~$e_1,\ldots,e_T$ 				&errors\\
	~~~~$\tau_1,\ldots,\tau_m$ 			&target counts for each bin\\        
        ~~~~$s$							&$\chi^2$ statistics\\
        \end{tabular}
        }
\end{center}    
\caption{Declarative statistics model for linear model fitting.}
\label{model:example_1_linear_regression_dscp_model}
\end{figure}

Another important step in classical linear model fitting is to determined confidence intervals for the statistical model parameters. As we will show next, this computation is naturally expressed in declarative statistics by slightly modifying the model in Fig. \ref{model:example_1_linear_regression_dscp_model} in such as way as to minimise (resp. maximise) the model parameter for which we aim to compute the confidence interval. Assuming, for instance, one aims to compute confidence intervals for the slope parameter $a$, we simply replace the original objective function with the following one: $\min$ $a$ (resp. $\max$ $a$). By employing this model, we obtain the $1-\alpha$ confidence intervals shown in Table \ref{model:example_1_linear_regression_ci}. 

\begin{table}
\centering
\begin{tabular}{lcc}
Parameter		&CI				\\
\hline
$a$			&	(-0.27,1.56)	\\
$b$			&	(-13.4,7.98)	\\
$\sigma$		&	(2.82,16.8)	\\
\end{tabular}
\caption{Confidence intervals.}
\label{model:example_1_linear_regression_ci}
\end{table}

\begin{figure}[h!]
\centering
\includegraphics{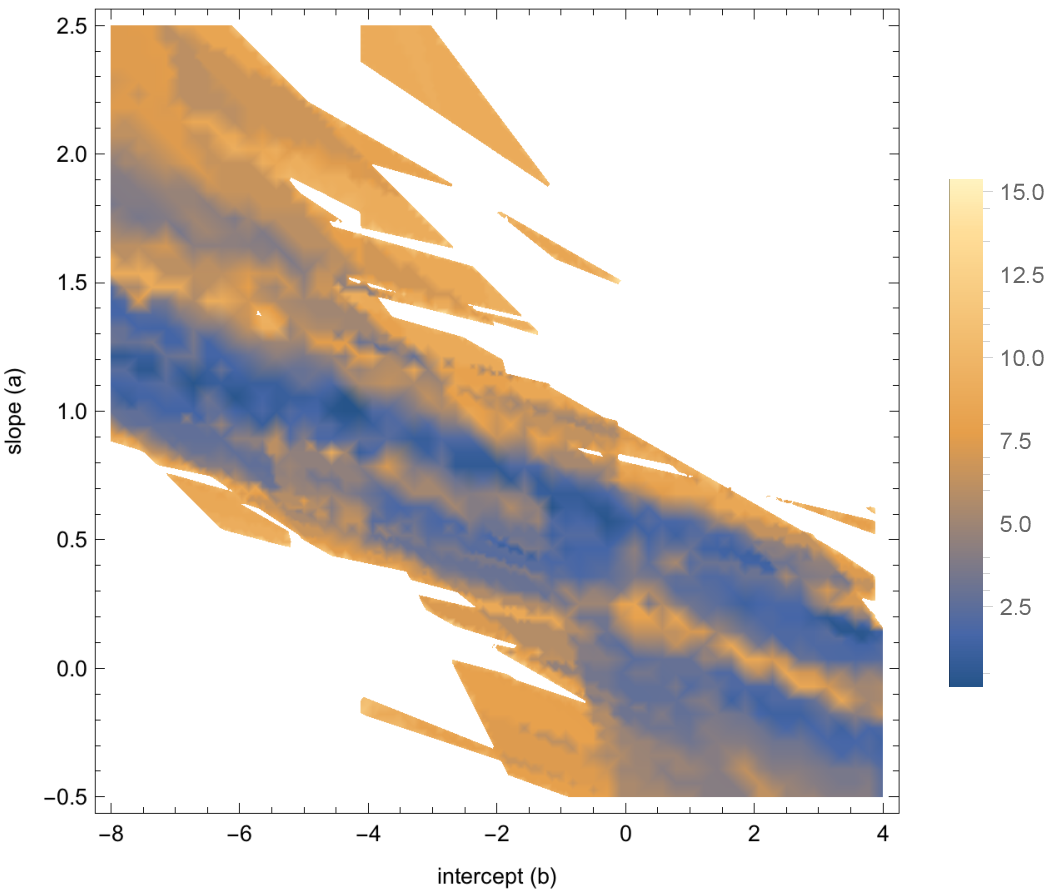}
\caption{Confidence region for slope ($a$) and intercept ($b$); white areas represent infeasible assignments, darker colours reflect a smaller (i.e. better) chi squared statistic.}
\label{fig:confidence_region_example_1}
\end{figure}

To show that the above confidence intervals are nominal, one can fix decision variables $a$, $b$ and $\sigma$ to their actual values: $a=1$, $b=-5$ and $\sigma=5$, which have been used to generate the random variates. If one then repeatedly generates new sets of $T=20$ random variates and solves the model in Fig. \ref{model:example_1_linear_regression_dscp_model} as a constraint satisfaction model, with probability $\alpha$ the model will be infeasible --- true parameter values will not be part of the solution space. In other words, the feasible region of our declarative statistics model is a representation of the confidence region associated with the statistical model parameters; a projection of this region over the slope ($a$) and intercept ($b$) with fixed $\sigma=5$ is given in Fig. \ref{fig:confidence_region_example_1}, white areas represent infeasible assignments, darker colours reflect a smaller (i.e. better) chi squared statistic. 

Before moving to the next application, it is worth remarking that the example here presented was purportedly simplistic. The confidence region obtained for slope and intercept are considerably larger than those obtained with other state-of-the-art linear regression techniques; this is caused by the adoption of a $\chi^2$ goodness of fit statistical constraint in the model, which reduces available degrees of freedom by binning data. In Appendix I, we further discuss this matter and show that declarative statistics can seamlessly capture these state-of-the-art models as well. The reader should keep in mind that the value of the approach we propose is mostly related to its flexibility and expressiveness: declarative statistics models can capture linear as well as non-linear relations and can deal with a broad range of probability distributions. Our next application deals with a more complex model.

\subsection{Time series analysis}\label{sec:time_series}

A stochastic process widely employed in time series analysis \citep{citeulike:9159964} is the autoregressive process of order $p$, AR($p$), which takes the following form
\begin{equation} 
x_t=c+\sum_{i=1}^p\beta_i x_{t-i} + \varepsilon_t
\end{equation}
where $c$ is a constant, $\alpha_i$ are the parameters of the model and $\varepsilon_t$ is a random noise --- typically a white noise.

\begin{figure}[h!]
\centering
\includegraphics{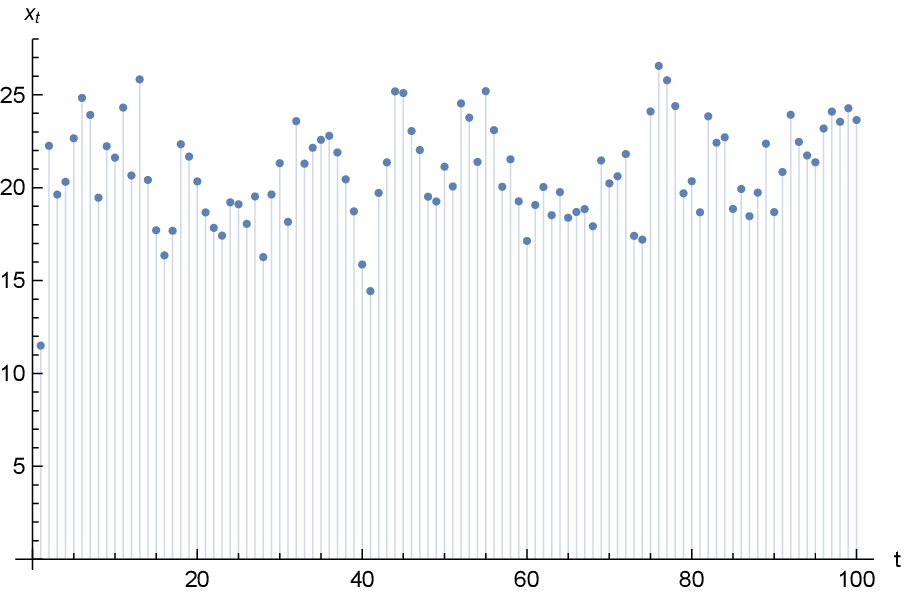}
\caption{Observations sampled from an autoregressive process of order 1.}
\label{fig:ar1_example}
\end{figure}

As a practical case, consider an AR(1) model with parameters $c=5$, $\beta=0.5$ and Poisson distributed random noise $\varepsilon_t$ with rate parameter $\lambda=5$; in Fig. \ref{fig:ar1_example} we plot 100 observations sampled from this stochastic process.

Given a set of observations like those in Fig. \ref{fig:ar1_example} one typically tries to determine what model parameter values are compatible with the data. This task is easily accomplished in declarative statistics. Consider a set of $T$ random variates $x_t$ for $t=1,\ldots,T$ and assume $x_0=0$; the model is shown in Fig. \ref{model:ar1_example_dscp_model}. 

\begin{figure}[h!]
\begin{center}
        \framebox{
        \begin{tabular}{ll}
        \mbox{\bf Objective:}\\
        ~~~~$\min$ $\beta$ (resp. $\max$ $\beta$)\\
        \\
        \mbox{\bf Constraints:}\\
	~~~~(1)~~$e_t=x_t-c-\beta x_{t-1}$						&for $t=1,\ldots,T$\\
	~~~~(2)~~$\tau_i=T (F_\lambda(b_{i+1})-F_\lambda(b_i))$	&for $i=1,\ldots,m$\\
        \multicolumn{2}{l}{
        ~~~~(3)~~$\chi^2_b(e;\tau;s)$}\\
        ~~~~(4)~~$s\leq Q$\\
        \\
        \mbox{\bf Parameters:} \\
        ~~~~$T$							&time periods\\
        ~~~~$x_1,\ldots,x_T$				&random variates\\
        ~~~~$m$							&number of bins\\
        ~~~~$b_1,\ldots,b_{m+1}$ 			&bin boundaries\\
        ~~~~$F_t$ 						&poisson cumulative distribution \\ 
        					   				&with rate $\lambda$ \\
	~~~~$Q$							&$1-\alpha$ quantile of the inverse $\chi^2$\\
        									&distribution with $m-1$ degrees\\ 
	        					   			&of freedom\\
        \\
        \mbox{\bf Decision variables:} \\
        ~~~~$\beta$						&AR(1) parameter\\
        ~~~~$c$							&AR(1) constant\\        
        ~~~~$\lambda$						&AR(1) poisson noise rate\\
        ~~~~$e_1,\ldots,e_T$ 				&errors\\
	~~~~$\tau_1,\ldots,\tau_m$ 			&target counts for each bin\\        
        ~~~~$s$							&chi squared statistics\\									
	\end{tabular}
        }
\end{center}    
\caption{Declarative statistics model for computation of confidence intervals of AR(1) parameters.}
\label{model:ar1_example_dscp_model}
\end{figure}

This model can be used to determine confidence intervals of AR(1) parameters; or, if in the objective function we minimise $s$, to find an optimal fit. In the previous example, assuming 15 unit bins $b_i\in\{0,\ldots,14\}$ and non-negative $c$ and $\beta$, the model returns the following confidence intervals at significance level $\alpha=0.05$: $c=(0,11.4)$; $\beta=(0, 0.79)$; and $\lambda=(1.34,15.0)$. If we minimize $s$, the fitted parameters are $c=6.28$; $\beta=0.467$; and $\lambda=5.06$.

Finally, consider two AR(1) processes of which we ignore parameters and the nature of the random noise. By leveraging the $\chi^2$ test of independence statistical constraint one obtains a very compact model that carries out the fit and allows confidence interval analysis. The model is shown in Fig. \ref{model:ar1_two_processes_example_dscp_model}. The idea is to impose a $\chi^2$ test of independence statistical constraint on the errors, which must be satisfied by any feasible assignment of $c^k$ and $\beta^k$, for $k=1,2$.

\begin{figure}[h!]
\begin{center}
        \framebox{
        \begin{tabular}{ll}
        \mbox{\bf Objective:}\\
        ~~~~$\min$ $s$\\
        \\
        \mbox{\bf Constraints:}\\
	~~~~(1)~~$e^1_t=x^1_t-c^1-\beta^1 x^1_{t-1}$				&for $t=1,\ldots,T$\\
	~~~~(2)~~$e^2_t=x^2_t-c^2-\beta^2 x^2_{t-1}$				&for $t=1,\ldots,T$\\
        \multicolumn{2}{l}{
        ~~~~(3)~~$\chi^2_{b,d}(e^1;e^2;s)$}\\
        ~~~~(4)~~$s\leq Q$\\
        \\
        \mbox{\bf Parameters:} \\
        ~~~~$T$							&time periods\\
        ~~~~$x^k_1,\ldots,x_T$				&random variates of process $k$\\
        ~~~~$m$							&number of bins\\
        ~~~~$b_1,\ldots,b_{m+1}$ 			&bin boundaries for noise 1\\
        ~~~~$d_1,\ldots,d_{n+1}$ 			&bin boundaries for noise 2\\        
	~~~~$Q$							&$1-\alpha$ quantile of the inverse $\chi^2$\\
        									&distribution with $(m-1)(n-1)$ degrees\\ 
	        					   			&of freedom\\
        \\
        \mbox{\bf Decision variables:} \\
        ~~~~$\beta^k$						&AR(1) parameter of process $k$\\
        ~~~~$c^k$						&AR(1) constant of process $k$\\        
        ~~~~$e^k_1,\ldots,e^k_T$ 			&errors of process $k$\\
        ~~~~$s$							&chi squared statistics\\									
	\end{tabular}
        }
\end{center}    
\caption{Declarative statistics model for the case of two AR(1) processes.}
\label{model:ar1_two_processes_example_dscp_model}
\end{figure}

\subsection{Comparing means of two or more samples}\label{sec:anova}


A classical application of statistics is to compare means of two or more samples.

If the aim is to compare the means of only two samples, this task is easily accomplished by mean of a Student's $t$ test. As discussed in Section \ref{sec:student}, in declarative statistics, this test is readily available in the form of a statistical constraint.

If the aim is to compare the means of three or more samples, one may execute a set of Student's $t$ test on each pair of samples while at the same time applying Bonferroni's correction to ensure that the familywise error rate complies with the chosen significance level for the test. 

However, more often, a one-way analysis of variance (ANOVA) \citep{Fisher1921} is performed to determine if any significant difference exists among the means.\footnote{For a description of one-way ANOVA see e.g. \citep{citeulike:14403808}.}

To focus attention, we consider the following example. We sampled $N=6$ independent realisations shown in Table \ref{model:example_2_anova_samples} from each of three normal random variables with mean $\mu_1=4$, $\mu_2=9$, and $\mu_3=10$. Since ANOVA operates under the assumption that population variances are equal, we shall assume the common standard deviation is $\sigma=3$.

\begin{table}
\centering
\begin{tabular}{ccc}
Group 1& Group 2& Group 3\\
\hline
 3.57329 & 9.83132 & 9.80335 \\
 6.5655 & 9.7379 & 8.79726 \\
 -2.06033 & 6.6339 & 13.6045 \\
 0.469477 & 8.20049 & 9.4932 \\
 3.05632 & 7.19737 & 8.50685 \\
 5.54063 & 9.19586 & 9.22433 \\
\end{tabular}
\caption{Samples considered in our example.}
\label{model:example_2_anova_samples}
\end{table}

\begin{figure}
\begin{center}
        \framebox{
        \begin{tabular}{ll}
        \mbox{\bf Constraints:}\\
        {\em ``between group'' mean squared differences}								\\
	~~~~(1)~~$\textsc{mean}(\bar{y}_i,\{O_{i,1},\ldots,O_{i,n}\})$		&for $i=1,\ldots,m$	\\
	~~~~(2)~~$\textsc{variance}(s_b,\{\bar{y}_1,\ldots,\bar{y}_m\})$					\\
	~~~~(3)~~$\bar{s}_b = n s_b / (m-1)$										\\
	\\
	{\em ``within group'' mean squared differences}									\\
	~~~~(4)~~$\textsc{variance}(s^i_w,\{O_{i,1},\ldots,O_{i,n}\})$		&for $i=1,\ldots,m$	\\
	~~~~(5)~~$\textsc{mean}(\bar{s}_w,\{s^1_w,\ldots,s^m_w\})$						\\
	\\
	{\em Fisher's $F$-ratio statistic}\\
	~~~~(6)~~$\bar{s}_b/\bar{s}_w\leq F^{-1}_{m-1,m(n-1)}(1-\alpha)$					\\
        \\     
        \mbox{\bf Parameters:} \\
        ~~~~$m$							&number of groups\\
        ~~~~$n$							&number of random variates \\
        									&within a group\\
        ~~~~$O_{i,j}$						&random variate $j$ in group $i$\\
        ~~~~$\alpha$						&significance level\\
        \\
        \mbox{\bf Decision variables:} \\
        ~~~~$\bar{y}_i$						&mean within group $i$\\
        ~~~~$n s_b$						&``between group'' sum of \\
        									& squared differences\\
	 ~~~~$\bar{s}_b$					&``between group'' mean \\
        									& squared differences\\								
	~~~~$\bar{s}_w$ 					&``within group'' mean \\
        									& squared differences\\								
        ~~~~$\bar{s}_b/\bar{s}_w$			&$F$-ratio statistic\\
        ~~~~$F_{a,b}$						&$F$-ratio distribution with \\
        									&$a$ degrees of freedom in the\\
									& in the numerator and $b$ in the\\
									& denominator. \\
        \end{tabular}
        }
\end{center}    
\caption{Declarative statistics model for one-way analysis of variance.}
\label{model:example_2_anova_dscp_model}
\end{figure}

ANOVA can be easily captured in declarative statistics. A model performing this well-known analysis is shown in Fig. \ref{model:example_2_anova_dscp_model}. The model is infeasible for the data shown in Table \ref{model:example_2_anova_samples} at signficance level $\alpha=0.05$. We recall that in declarative statistics, an infeasibility signifies rejection of the null hypothesis. In ANOVA the null hypothesis is that the means of the normal distributions that generated the three group realisations are equal, we therefore conclude that there is evidence that the expected values in the three groups differ at the prescribed significance level. 

Unfortunately, the model in Fig. \ref{model:example_2_anova_dscp_model} does not provide much insight on what group(s) averages may be equal or different. A standard ANOVA table for the example can be obtained from standard software packages such as Mathematica\textsuperscript{TM} (Table \ref{table:example_2_anova_table}). In line with our findings, the table reveals that the $p$-value --- the probability of finding the observed, or more extreme, results when the null hypothesis is true --- is less than the prescribed significance level. Typically a so-called ``post-hoc'' analysis is necessary to further investigate what group mean(s) caused the null hypothesis rejection. In the case of our numerical example a post-hoc Tukey's test would reveal that the mean of Group 1 differ at the prescribed significance level from that of Group 2 and Group 3.

\begin{table}
\centering
\begin{tabular}{lllllll}
\multirow{4}{*}{ANOVA}&
			&DF	&SumOfSq	&MeanSq	&$F$-Ratio	&$p$-value\\
&Model		&2	&166.387	&83.1935& 16.0089& 0.000189984\\
&Error		&15	&77.9505	&5.1967\\
&Total		&17	&244.337\\
\hline
\multirow{4}{*}{Means}&
Overall	&7.07618\\
&Group 1	&2.85748\\
&Group 2	&8.46614\\
&Group 3	&9.90492
\end{tabular}
\caption{One-way analysis of variance table; DF: degrees of freedom; SumOfSq: sum of squared differences; MeanSq: mean squared differences.}
\label{table:example_2_anova_table}
\end{table}

Instead of relying on a two-step procedure involving post-hoc analysis, we will illustrate next a simpler solution that leverages the Hotelling's $t^2$ statistical constraint previously introduced. The compact model shown in Fig. \ref{model:example_2_hotelling_dscp_model} directly captures the problem of estimating the mean of a multivariate normal distribution from a set of samples. By minimizing Hotelling's $t^2$ statistic for the dataset previously analysed we obtain the following estimated group means: $\mu_1=2.70$, $\mu_2=8.38$, $\mu_3=9.93$, which are in line with those previously obtained via ANOVA. 

\begin{figure}[h!]
\begin{center}
        \framebox{
        \begin{tabular}{ll}
        \mbox{\bf Objective:}\\
        ~~~~$\min$ $s$\\
        \\
        \mbox{\bf Constraints:}\\
        \multicolumn{2}{l}{
        ~~~~(1)~~$t^2(\langle O\rangle;\mu;s)$}\\
        ~~~~(2)~~$s\leq Q$\\
        \\     
        \mbox{\bf Parameters:} \\
        ~~~~$m$							&number of groups\\
        ~~~~$n$							&number of random variates \\
        									&within a group\\
        ~~~~$O_{i,j}$						&random variate $j$ in group $i$\\
        ~~~~$Q$							&$1-\alpha$ quantile of the inverse \\
        									&Hotelling's $T^2_{m,n-1}$ distribution\\
        \\
        \mbox{\bf Decision variables:} \\
        ~~~~$\mu_i$						&mean of group $i$\\
        ~~~~$s$							&Hotelling's $t^2$ statistic\\
        \end{tabular}
        }
\end{center}    
\caption{Declarative statistics model for multivariate normal parameter fitting.}
\label{model:example_2_hotelling_dscp_model}
\end{figure}

By introducing small modifications to the model, it is simple to push the analysis further. If we want to investigate whether the assumption that $\mu_i=\mu_j$ for all $i,j=1,\ldots,m$ and $j>i$ holds, i.e. all group means are equal, we can simply post these constraints in the model. If we do so in our numerical example, the model becomes infeasible. Similar to what we discovered via ANOVA, the null hypothesis that group means are equal is rejected at signficance level $\alpha$. If, however, we only post constraint $\mu_2=\mu_3$, the model remains feasible; this means we fail to reject the hypothesis that means of groups 2 and 3 are equal; this is in line with the results previously obtained via a post hoc Tukey's test. If we want to derive confidence intervals of $\mu_i$, we simply have to replace the objective function with $\min$ $\mu_i$ (resp. $\max$ $\mu_i$) to determine lower (resp. upper) confidence bounds. It is worth noting that these confidence intervals change if we post additional constraints in the model. For instance, originally, the confidence interval for $\mu_3$ is $(5.91, 13.4)$; if however we post constraint $\mu_2=\mu_3$, the interval shrinks to (7.51, 10.5). This shows the flexibility, expressiveness, and power of our new framework. 

\subsection{Confidence interval analysis for multinomial proportions}\label{sec:multinomial}

As discussed in the previous section Hotelling's $t^2$ statistic can be used to carry out hypothesis testing on multivariate distributions (see \cite{citeulike:14214028}, chap. 5). In this section we shall concentrate on the multinomial distribution (see \cite{citeulike:14214023}, section 3). 

Consider a multinomial distribution with event probabilities $p_1,\ldots,p_k$, where $k$ is the number of categories, and $N$ trials. Let $x_1,\ldots,x_n$ be $n$ i.i.d. random variates and $c_1,\ldots,c_k$ be associated observed cell counts in a sample size of $N=\sum_{i=1}^k c_i$. The problem of determining simultaneous confidence intervals for $p_1,\ldots,p_k$ was developed in \citep{citeulike:14214031,citeulike:14214026,10.2307/1266673}.

The maximum likelihood estimators of $p_j$ are $\hat{p}_j=c_j/N$, $j=1,\ldots,k$. The random vector $\hat{p}\equiv(\hat{p}_1,\ldots,\hat{p}_k)$ is asymptotically distributed according to a multivariate normal distribution with mean vector $p\equiv(p_1,\ldots,p_k)$ and covariance matrix $\Sigma$ with elements $\sigma_{jj}=p_j(1-p_j)$ and $\sigma_{ij}=-p_ip_j$ for $i\neq j$. \cite{citeulike:14214026} discussed confidence intervals based on the quadratic form
\[N(\hat{p}-p)'\Sigma^{-1}(\hat{p}-p)\]
which is asymptotically distributed as a $\chi^2$ distribution with $k-1$ degrees of freedom. Let $1-\alpha$ be the desired confidence level; confidence intervals are obtained, for $j=1,\ldots,k$, as the two solutions of equation
\[N\frac{(\hat{p}_j-p_j)^2}{p_j(1-p_j)}=F^{-1}_{\chi^2_{k-1}}(1-\alpha).\]

Alternatively, in declarative statistics, the quadratic form presented above can be easily captured via Hotelling's $\chi^2$ statistic constraint, which we have previously introduced. The model is shown in Fig. \ref{model:chi2_test_multinomial} and can be expressed in a very compact way via a single statistical constraint. As before, the feasible region of this model covers ``true'' multinomial event probabilities in line with the prescribed confidence level $1-\alpha$.

A faster model, which however leads to more conservative confidence intervals, can be obtained by replacing constraint (1) with Hotelling's $t^2$ statistical constraint, $t^2(\langle X\rangle;p;s)$, in which the covariance matrix $\Sigma$ is assumed unknown and replaced by its sample value. 

In both cases, the random variates matrix should be appropriately trimmered according to the discussion in \citep{citeulike:14403078}, to make sure the variance-covariance matrix is nonsingular. In simple terms, one should simply drop one of the columns associated with a category that is not being estimated. 

\begin{figure}[h!]
\begin{center}
        \framebox{
        \begin{tabular}{ll}
        \mbox{\bf Objective:}\\
        ~~~~$\min$ $p_i$ (resp. $\max$ $p_i$)\\
        \\
        \mbox{\bf Constraints:}\\
        \multicolumn{2}{l}{
        ~~~~(1)~~$\chi_\alpha^2(\langle X\rangle;p;\Sigma)$}\\
        \\
        \mbox{\bf Decision variables:} \\
        ~~~~$X$ 							&random variates matrix\\
        ~~~~$\Sigma$ 						&covariance matrix with elements \\
        									&$\sigma_{jj}=p_j(1-p_j)$ and \\
									&$\sigma_{ij}=-p_ip_j$ for $i\neq j$ \\
        ~~~~$p_1,\ldots,p_k$ 				&multinomial event probabilities\\
        \end{tabular}
        }
    \caption{A declarative statistics model for computing confidence intervals for multinomial proportions. }
    \label{model:chi2_test_multinomial}
\end{center}    
\end{figure}

As a practical case, we consider $N=10$ i.i.d. observations drawn from a multinomial with event probability vector $p=\{0.3,0.3,0.4\}$. The random variates matrix is shown in Table \ref{tab:example_chi2_test_multinomial}, the associated cell counts are $c=\{3,5,2\}$. We set the target significance for the score test $\alpha=0.1$ (i.e. a confidence level $1-\alpha=0.9$); in Table \ref{tab:confidence_intervals} we compare confidence intervals obtained using Quesenberry and Hurst's closed form expressions and intervals obtained as solutions of our model (Fig. \ref{model:chi2_test_multinomial}) based on Hotelling's $t^2$ statistic constraint. Since Quesenberry and Hurst's closed form expressions rely on a heuristic decomposition of the problem, they are necessarily more conservative than our model based on Hotelling's $t^2$ statistic.

\begin{table}
\centering
\begin{tabular}{ccccc}
\{0,1,0\}&\{0,1,0\}&\{0,0,1\}&\{1,0,0\}&\{0,1,0\}\\
\{0,1,0\}&\{0,1,0\}&\{1,0,0\}&\{0,0,1\}&\{1,0,0\}
\end{tabular}
\caption{Random variates for our numerical example}
\label{tab:example_chi2_test_multinomial}
\end{table}


\begin{table}[h!]
\centering
\begin{tabular}{l|rr}
								&Quesenberry and Hurst's					&Hotelling's $t^2$ statistical constraint\\
\hline							
$(p^{\text{lb}}_1,p^{\text{ub}}_1)$		&(0.0981, 0.6280)	&(0.0731,0.5267)\\
$(p^{\text{lb}}_2,p^{\text{ub}}_2)$		&(0.2192, 0.7808)	&(0.2526,0.7474)\\
$(p^{\text{lb}}_3,p^{\text{ub}}_3)$		&(0.0509, 0.5383)	&(0.0020,0.3978)
\end{tabular}
\caption{Confidence intervals for our numerical example}
\label{tab:confidence_intervals}
\end{table}

\section{Related works}\label{sec:related_works}

``Declarative statistics,'' i.e. the development of declarative languages for statistical inference, has been recently advocated in \citep[][Section 6.2]{citeulike:14409087}. As the authors remark, ``statistical inference is too complex and too subtle to be left to occasional users;'' and several articles have pointed out incorrect use of statistical methods in machine learning and data mining \cite[see e.g.][]{Dietterich:1998:AST:303222.303237,Demsar:2006:SCC:1248547.1248548}. Unfortunately, ``the current state of the art is such that correct conclusions can only be drawn if users select the right statistical test (if one exists), perform it correctly, and interpret the results correctly.'' 

A recent stream of research aimed to introduce declarative frameworks for specifying statistical models in database systems.
\cite{Vanwinckelen2013} discussed the development of an SQL-like language for statistical inference on the top of a database. Their work presents preliminary ideas and a conceptual roadmap that is closely aligned to the discussion in our work. \cite{DBLP:journals/corr/BaranyCKOV14,barany_et_al:LIPIcs:2016:5776} blend Probabilistic Programming (PP) \citep{Goodman:2013:PPP:2429069.2429117} and Datalog \citep{abiteboul:hal-00923265} with the aim of specifying statistical models, and establishing a declarative PP paradigm over databases. PP is an extension of probabilistic graphical models \citep{Koller:2009:PGM:1795555}, also known as Bayesian networks. Bayesian networks represent conditional probability relationships between a set of variables in the form of a directed acyclic graph. Bayesian inference carried out in PP --- typically performed by inference engines based on variants of Markov Chain Monte Carlo, most notably Metropolis-Hastings --- is different from the one outlined in our work, which is based classical hypothesis testing techniques. Moreover, as \cite{citeulike:14409087} remark, ``It is clear that more research is needed before such approaches will be widely useful in practice, but the potential impact of such research is large.'' Finally, another relevant challenge in Bayesian networks is determining which graph best explains observed data. For this NP-hard problem,  \cite{citeulike:14410889} recently discussed an integer linear programming approach.

In line with the research agenda discussed in the previous paragraphs, our work also aims to develop a declarative frameworks for specifying statistical models, but rather than focusing on database systems and Bayesian networks, we target constraint programming systems and established techniques in statistical hypothesis testing. We build upon recent Choco \citep{choco} extensions for combining finite and continuous solvers \citep{tricks2013}; in particular, these extensions delegate constraint processing over continuous domains to a separate library for constraint processing over real numbers: Ibex. In this sense, we see our work as a development in the context of uncertainty modeling in constraint programming.

Comprehensive surveys on uncertainty modeling in constraint programming can be found in \citep{BrownMiguel06,citeulike:8353640}. Two notable frameworks for uncertainty modeling in constraint programming are probabilistic constraint programming \citep{DBLP:conf/uai/FargierLMS95} and stochastic constraint programming \citep{DBLP:conf/ecai/Walsh02}. Probabilistic constraint programming is an extension of constraint programming  mainly concerned in expressing uncertainty on the values of problem parameters as a probability distribution, which is assumed to be available. Stochastic constraint programming works with decision and random variables over a set of decision stages; also in this case random variable distributions are assumed to be known. Our framework operates under the assumption that distribution of random variables --- i.e. of the {\em sample}, to use the terminology introduced in Section \ref{sec:statistical_constraints} --- is only partially specified (parametric statistical constraints) or not specified at all (non-parametric statistical constraints); furthermore, we do not model these random variables explicitly, we model instead random variates as decision variables. 

Although the present work does not focus on the stochastic constraint programming framework, it nevertheless complements \citep{citeulike:13269658,citeulike:13708770}, in which statistical inference is applied in the context of stochastic constraint satisfaction/optimisation to identify approximate solutions featuring given statistical properties. The philosophy underpinning both works is similar: both confidence-based reasoning and declarative statistics recognize that --- when limited information is available to perform an inference --- the decision maker should express a solution and/or its value in terms of a confidence region, not in terms of a point estimate. This approach has been recently advocated in a number of works \citep[see e.g.][]{citeulike:14023064,citeulike:14023061,citeulike:13384427}. 

We contributed to the constraint programming literature on counting and value constraints by defining the \textsc{bin\_counts} --- directly related to the \textsc{frequency} constraint discussed in \citep{citeulike:14393881} --- and the \textsc{contingency} constraint; and by proposing decompositions for them. Finally, we also introduced the \textsc{matrix\_inversion} constraint and proposed a decomposition for it; to the best of out knowledge, the literature on constraints capturing matrix calculus is limited and more research is needed in this direction, given that these constraints are essential for statistical modeling.

Constraints capturing statistics such as mean and standard deviation have been long known and investigated in the constraint programming literature, see e.g. \citep{citeulike:13171963,citeulike:14181517,monette2013parametric,bcspread,citeulike:14181514,bessiere:lirmm-01374715,loong2016}. As we outlined in previous sections, these studies are particularly relevant in the context of our work. We contribute to this stream of literature by introducing new constraints capturing statistics, such as the \textsc{standard\_error} and the \textsc{covariance} constraints, which are instrumental in the development of new statistical constraints. The development of more efficient filtering algorithms for these constraints clearly represents an important future research direction. 

Finally, since the publication of \citep{rossi2014}, a number of follow up studies have appeared \citep{Pachet:2015:GNS:2832581.2832596,sony2016,citeulike:14393881} in relation to statistical constraints, and to constraints capturing randomness \citep{citeulike:13771650}. This stream of literature testifies the relevance of developing declarative frameworks for statistical inference in constraint programming.

\section{Conclusions}\label{sec:conclusions}

In this work we introduced a suite of declarative modelling tools for statistical analysis in constraint programming. Our contribution to the literature is threefold. First, we introduced two counting constraints --- the \textsc{bin\_counts} and the \textsc{contingency} constraints --- a novel \textsc{matrix\_inversion} constraint, and associated decompositions. Second, by leveraging these new constraints as well as known constraints capturing statistics, such as \textsc{mean} and \textsc{variance}, we introduced a range of new statistical constraints: the $\chi^2$ goodness of fit, the 
$\chi^2$ test of independence, Fisher's ratio, and Hotelling's $t^2$ statistical constraints. Third, we discussed applications of these constraints in the context of a number of classical domains encountered in statistics: linear model fitting, time series analysis, sample mean comparison, and computation of confidence intervals for the multinomial distribution. We contrasted our approach against established methods and illustrated similarities and differences. In the case of linear model fitting, our method can mimic as well as generalise functionalities that are available in existing tools for statistical analysis, e.g. \texttt{LinearModelFit} module of Mathematica\textsuperscript{TM}. In the case of sample mean comparison, our method produces a model that is simpler and more intuitive than an ANOVA table. In the case of the computation of confidence intervals for multinomial proportions, our method produces a very compact and intuitive model that leads to confidence intervals that are competitive against established approaches. Our models are generally compact and should be intelligible both to mathematical programmers and statisticians. The framework is currently implemented as a Choco extension, and thus it is aimed at an audience possessing coding skills. Future research should focus on the design and implementation of a full fledged algebraic modeling language that will make the framework accessible to a broader audience.

\section*{Acknowledgments}\label{sec:conclusions}
R. Rossi is supported by Volvo Construction Climate Challenge fund and by EPSRC Industrial Case 2017. 

\newpage

\section*{Appendix I}


In this section we briefly illustrate two alternative declarative statistics models for linear model fitting.

As before, consider a set of $T$ random variates $v_t$ for $t=1,\ldots,T$. We assume that variates have been generated according to 
\[v_t=a t+b+e_t\]
where $e_t$ is a random variable with cumulative distribution function $F_\sigma$. Without loss of generality and in line with established practices in statistics, we shall assume that $F_\sigma$ is normal with mean zero and standard deviation $\sigma$.

We distinguish two cases in the following discussion: in the first case, $\sigma$ is assumed to be known; in the second, it has to be estimated from the data. 

\begin{figure}[h!]
\begin{center}
        \framebox{
        \begin{tabular}{ll}
        \mbox{\bf Constraints:}\\
	~~~~(1)~~$e_t=v_t-(a t+b)$			&for $t=1,\ldots,T$\\
	~~~~(2)~~$\chi_\alpha^2(e;\mu;\sigma)$		\\
        \\     
        \mbox{\bf Parameters:} \\
        ~~~~$T$							&time periods\\
        ~~~~$v_1,\ldots,v_T$				&random variates\\
        ~~~~$\mu$						&normal mean, fixed and set to 0\\        
        ~~~~$\sigma$						&normal standard deviation\\        
        \\
        \mbox{\bf Decision variables:} \\
        ~~~~$a$							&fitted model slope\\
        ~~~~$b$							&fitted model intercept\\        
        ~~~~$e_1,\ldots,e_T$ 				&errors
        \end{tabular}
        }
\end{center}    
\caption{Declarative statistics model for linear model fitting: known $\sigma$.}
\label{model:appendix_1_linear_regression_dscp_model_known_sigma}
\end{figure}

\begin{figure}[h!]
\centering
\includegraphics[width=0.6\columnwidth]{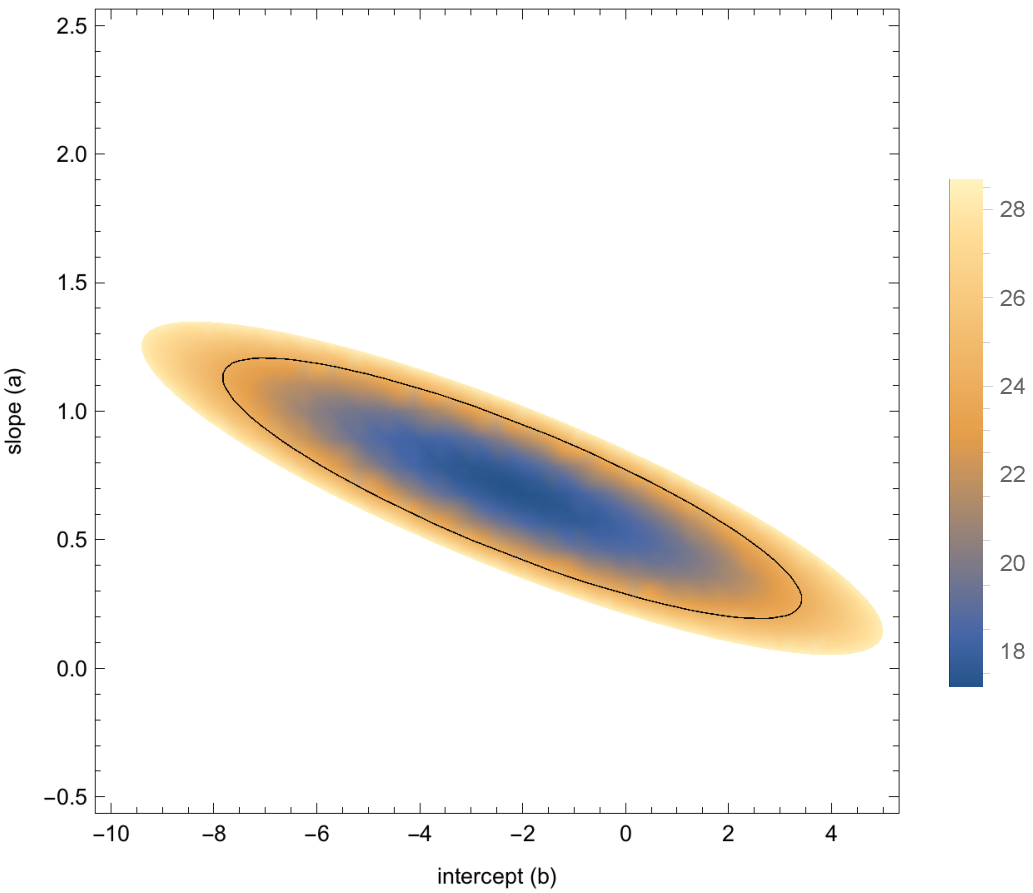}
\caption{Confidence region for slope ($a$) and intercept ($b$) under the assumption that $\sigma$ is known; the colour gradient reflects the value of the $\chi^2$ statistics.}
\label{fig:appendix_I_chi_square}
\end{figure}

To address the first case, we slightly modify the model presented in Section \ref{sec:regression} as shown in Fig. \ref{model:appendix_1_linear_regression_dscp_model_known_sigma}. Rather than relying on a $\chi^2$ goodness of fit statistical constraint, we rely on a special univariate case of Hotelling's $\chi^2$ test (constraint 2). For the example presented in Section \ref{sec:regression}, when $\alpha=0.95$, the feasible region of the model is shown in Fig. \ref{fig:appendix_I_chi_square}. The ellipsoid shown in black, has been obtained via the \texttt{LinearModelFit} module of Mathematica\textsuperscript{TM} (ParameterConfidenceRegion, ConfidenceLevel: 0.95); this ellipsoid is visibly smaller than the region obtained via our model. This is due to the fact that Mathematica fits three model parameters (slope: $a$, intercept: $b$, noise standard deviation: $\sigma$) from the data, and thus accounts for the respective reduction in degrees of freedom. To obtain the same region, we can replace constraint 2 with 
\[
\begin{array}{l}
\mathrm{~~~~(2)~~}\chi^2(e;\mu;\sigma;s)\\
\mathrm{~~~~(3)~~}s\leq Q
\end{array}
\]
where $s$ is the $\chi^2$ statistics and $Q$ is the $1-\alpha$ quantile of the inverse $\chi^2$ distribution with $T-3$ degrees of freedom. However, this would lead to a reduced --- non-nominal --- coverage probability of 0.9 for true parameters values in a repeated experiments setting. To understand the rationale for this, we must underscore the fact that our model does not fit parameters, it provides a confidence region for them, thus in our model the correct number of degrees of freedom that attains nominal coverage probability is $T$. In other words, if we repeatedly generate new sets of observations like those in Table \ref{tab:example_1_random_variates} and, for each of these new sets of random variates $v_1,\ldots,v_T$ we solve the model in Fig. \ref{model:appendix_1_linear_regression_dscp_model_known_sigma}, with probability $1-\alpha$ the feasible region of the model will contain the ``true'' values of slope $a$, intercept $b$, and noise standard deviation $\sigma$ that generated the time series. Although, as seen, it is possible to obtain a single ``fitted'' assignment for $a$, $b$, and $\sigma$ --- for instance by minimizing $s$ --- the main purpose of the model is not to return this assignment, but to provide a black-box representation for the confidence region of problem parameters, which can be manipulated and queried by a decision maker via a high level modeling framework.

\begin{figure}[h!]
\begin{center}
        \framebox{
        \begin{tabular}{ll}
        \mbox{\bf Constraints:}\\
	~~~~(1)~~$e_t=v_t-(a t+b)$			&for $t=1,\ldots,T$\\
	~~~~(2)~~$t_\alpha^2(e;\mu)$\\
       \\     
        \mbox{\bf Parameters:} \\
        ~~~~$T$							&time periods\\
        ~~~~$v_1,\ldots,v_T$				&random variates\\
        \\
        \mbox{\bf Decision variables:} \\
        ~~~~$a$							&fitted model slope\\
        ~~~~$b$							&fitted model intercept\\        
        ~~~~$e_1,\ldots,e_T$ 				&errors
        \end{tabular}
        }
\end{center}    
\caption{Declarative statistics model for linear model fitting: unknown $\sigma$.}
\label{model:appendix_1_linear_regression_dscp_model_unknown_sigma}
\end{figure}

\begin{figure}[h!]
\centering
\includegraphics[width=0.6\columnwidth]{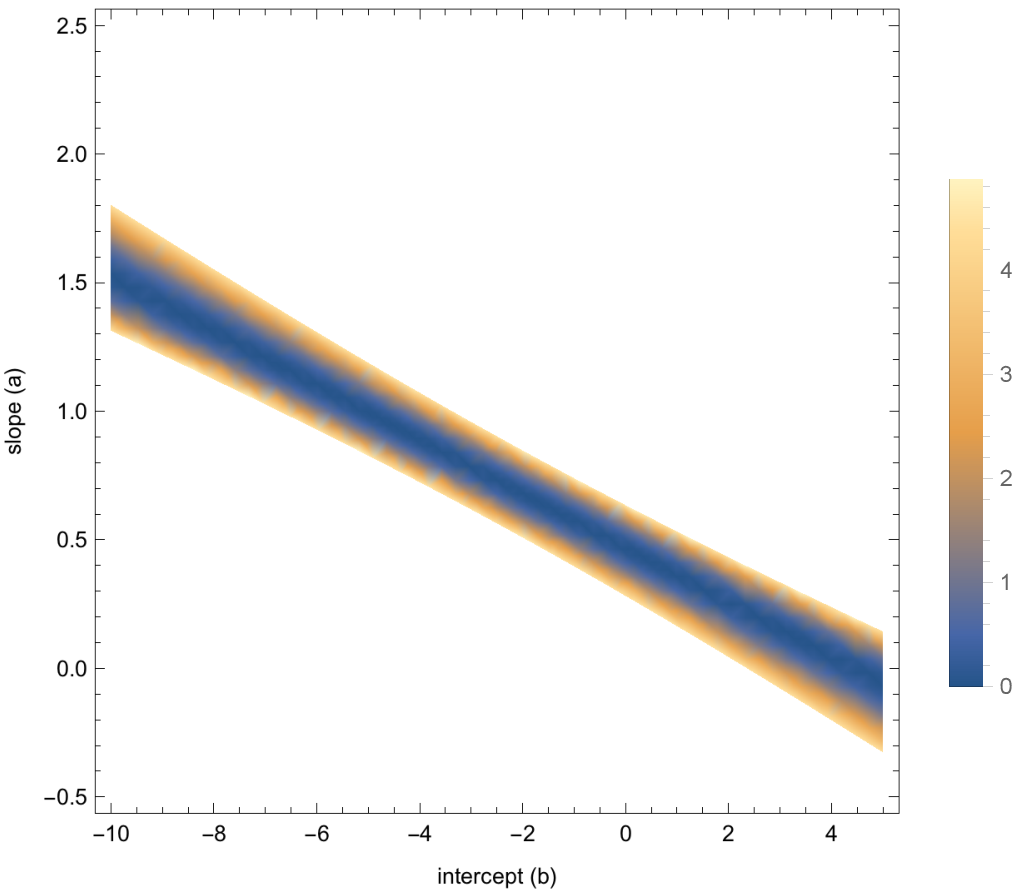}
\caption{Confidence region for slope ($a$) and intercept ($b$) under the assumption that $\sigma$ is estimated from the data; the colour gradient reflects the value of the $t^2$ statistics.}
\label{fig:appendix_I_t_square}
\end{figure}

The second case, in which $\sigma$ is estimated from the data, can be addressed via the model presented in Fig. \ref{model:appendix_1_linear_regression_dscp_model_unknown_sigma}. Constraint (2) is a special univariate case of Hotelling's $t^2$ statistical constraint. For the example presented in Section \ref{sec:regression}, when $\alpha=0.95$, the feasible region of the model is shown in Fig. \ref{fig:appendix_I_t_square}. As it is possible to observe, this revised assumption considerably modifies the structure of the feasible region of our model and, consequently, the confidence region for our model parameters.

\bibliographystyle{plainnat}
\bibliography{declarative_statistics}

\end{document}